\documentclass[review,11pt]{elsarticle}
\usepackage[utf8]{inputenc}
\usepackage[T1]{fontenc}
\usepackage{lineno,hyperref}
\usepackage{array}
\usepackage{amssymb}
\usepackage{enumerate}
\usepackage{makecell}
\usepackage{mathtools}
\usepackage{array}
\usepackage{multirow}
\usepackage{ulem}
\usepackage{setspace}
\usepackage{color}
\usepackage{booktabs}
\usepackage{diagbox}
\usepackage{enumerate}
\usepackage{graphicx}
\usepackage{subfigure}
 \usepackage{tikz,mathpazo} 
\usepackage{indentfirst}
\usepackage{soul}
\usepackage{amsmath}
\usepackage[justification=centering]{caption}
\usepackage{tabularx}
\usepackage{algorithm}
\usepackage{algorithmicx}
\usepackage{algpseudocode}
\usepackage{adjustbox}
\usepackage{geometry}
\usetikzlibrary{shapes.geometric, arrows}
\modulolinenumbers[5]

\soulregister{\cite}7
\soulregister{\ref}7

\newcommand{\circledsmall}[1]{\lower.7ex\hbox{\tikz\draw (0pt, 0pt)%
    circle (.5em) node {\makebox[0.1em][c]{\small#1}};}}

\newcommand{\circledtiny}[1]{\lower.7ex\hbox{\tikz\draw (0pt, 0pt)%
    circle (.3em) node {\makebox[0.1em][c]{\tiny #1}};}}

\journal{Journal of \LaTeX\ Templates}

\begin{document}

\begin{frontmatter}

\title{A Matrix-based Distance of Pythagorean Fuzzy Set and its Application in Medical Diagnosis}

    \author[label1,label2]{Yuanpeng He\corref{cor1}}
    \author[label3]{Lijian Li}
    \author[label3]{Tianxiang Zhan}
    
    \affiliation[label1]{organization={Key Laboratory of High Confidence Software Technologies (Peking University), Ministry of Education},
    	city={Beijing},
    	postcode={100871},
    	country={China}}
    
    \affiliation[label2]{organization={School of Computer Science, Peking University},
    	city={Beijing},
    	postcode={100871},
    	country={China}}

    \affiliation[label3]{organization={Department of Computer and Information Science, University of Macau},
    	city={Macau},
    	postcode={999078},
    	country={China}}
    	
%
%

    \cortext[cor1]{Corresponding author: Yuanpeng He is with Key Laboratory of High Confidence Software Technologies (Peking University), Ministry of Education, Beijing, China; School of Computer Science, Peking University, Beijing, China. E-mail address: heyuanpeng@stu.pku.edu and heyuanpengpku@gmail.com}

\begin{abstract}
The pythagorean fuzzy set (PFS) developed based on intuitionistic fuzzy set, is more efficient in elaborating and processing uncertainties in indeterminate situations. How to measure the distance between two PFSs is still an open issue. Many kinds of methods have been proposed to address this problem. However, not all of existing methods can accurately display differences among PFSs and satisfy the property of similarity. Some methods neglect the relationship among three variables of PFS. To address the problems, a new method of distance measure is proposed which meets the requirements of axiom of distance measurement and is developed by generalizing the concept of score function in a matrix form to better measure parameters difference of PFS. In detail, our proposal takes into account the hesitance index and allocates its weight between membership and non-membership in a rational manner, enhancing the significance of both membership and non-membership in shaping the distances between PFSs. Some numerical examples are offered to show that the proposed method can avoid producing counter-intuitive results and is more reasonable than other previous methods. Besides, the proposed method is applied to real-world environments of application and compared with previous methods to demonstrate its superiority and efficiency.
\end{abstract}


\end{frontmatter}

\section{Introduction}


There are lots of uncertainty existing in the real world, and how to measure the level of uncertainty has attracted incremental attention and interests from researches all around the world \cite{Deng2020ScienceChina,DBLP:journals/tfs/WuZYXLW24,DBLP:journals/tit/ChenL24}. Therefore, many meaningful and related theories have been proposed to be applied into practical applications, such as evidence theory \cite{DBLP:journals/isci/HeD23,DBLP:journals/inffus/LiuLXS23,DBLP:journals/tcyb/HuangLD23,DBLP:journals/tsmc/CuiD23}, belief function \cite{DBLP:journals/tcyb/ZhouDY24,DBLP:journals/apin/HeX22,DBLP:journals/inffus/KangZ24,DBLP:journals/cam/HeD22}, and structure \cite{DBLP:journals/tsmc/HuangGSZJY23,Yager2019,DBLP:journals/tfs/BhowalSYGS23}, entropy theory \cite{DBLP:journals/tit/MasihuddinM24,DBLP:journals/tie/Novak24,DBLP:journals/tit/WuLXH24,DBLP:journals/soco/HeD23}, $Z$ numbers \cite{Kang2019,dehshiri2024evaluation,li2020newuncertainty,mandal2024failure} and $D$ numbers \cite{Liu2019b,sotoudeh2024state,xiao2019multiple,Deng2019}, which play important roles in all walks of life. Except for the theories mentioned above, fuzzy sets \cite{Zadeh1965,Zadeh1979}, in seeking for useful information among uncertainties, becomes a key component of pattern recognition and decision making \cite{DBLP:journals/isci/RaniCM24,DBLP:journals/tfs/0001DX023,he2022ordinal}. 

Obviously, when judging actual situations in real world, things are getting more complex, which indicates that a single value can not reflect the essence of certain objects. Therefore, a further concept based on the fuzzy set is proposed such as intuitionistic fuzzy set (IFS). It is developed by Atanassov \cite{Atanassov1999} and defined to contain $3$ properties, namely membership, non-membership and hesitance, IFS is considered more efficient in handling ambiguity. It is further redesigned to satisfy different demands and be applicable in more complex situations, such as interval-valued intuitionistic fuzzy set \cite{DBLP:journals/isci/JebadassB24,DBLP:journals/eswa/DongW24} and quantum decision \cite{DBLP:journals/inffus/TiwariZQM24,DBLP:journals/eswa/MartinoS22}. After that, a new extension is invented by Yager \cite{Yager2014a} called pythagorean fuzzy set (PFS). It is a quadratic form of the original fuzzy set, which means that the new modality of fuzzy set has a larger range of the change of variables and therefore has more potential to express uncertainties. PFS is also extended into other forms, such as interval-valued pythagorean fuzzy set \cite{DBLP:journals/ijis/LvZZLK22} and applied in real-world problems \cite{DBLP:journals/eswa/AkramLA22,DBLP:journals/tfs/EjegwaWFZT22,DBLP:journals/asc/SarkarCB23}.

For any kind of fuzzy sets, how to measure the differences among them is an unavoidable problem of practical significance. In order to express the distances properly, many methods of distance measure have been proposed and some of them have superb performance in classification problems. The most widely used method of measuring distances between IFSs are the Hamming distance \cite{SZMIDT2000505}, Euclidean distance \cite{SZMIDT2000505}, the Hausdorff metric \cite{GRZEGORZEWSKI2004319}, Own's distance \cite{article}, De et al.'s distance \cite{DE2001209}, Szmidt et al.'s distance \cite{10.1007/3-540-45718-6_30}, Wei et al.'s distance \cite{WEI20114273}, Mondal et al's distance \cite{article1} and others. However, not all of the methods can produce intuitive results under any circumstances and may be obviously conflicting with data given. Because PFS is a generalization of IFS, the method of measuring distance between IFSs can be rationally extended into the forms which can measure distance between PFSs. But similarly, the challenges that exist in measuring IFS are also present when measuring PFS.

Therefore, to address the potential counter-intuitive and irrational results that existing work may introduce into the measurement of PFS, a novel method of distance measure of PFSs is proposed in this paper. Different distances have different emphases, and there are varying advantages in measuring the degree of differences in PFSs. Due to the hesitation index being a feature of PFSs that describes the states of "membership or non-membership," unlike the first two parameters of PFSs, the information carried by the hesitation is uncertain. As a result, when comparing the differences between PFSs, the three parameters cannot be considered separately and assigned equal importance. In detail, the method transforms the elements contained in the PFSs into a form of vector to maximize the role every element plays in generating distances taking the concept of generalized score function. The newly defined matrix is utilized to strengthen the membership and non-membership and the index of hesitance is also considered but its influence in distance measurement is more limited than the former two. Compared with the results produced by other methods, the ones from the proposed method is obviously more accurate and conforms to reality, which indicates it may be more efficient in practical applications. 

The rest of the paper is organized as follows. In the section of preliminaries, some relative concepts are briefly introduced. Besides, details are clearly provided and some simple cases are utilized to introduce the proposed method more straightforward in the section of proposed method. For the next section, the correctness and validity of the proposed method in distance are verified. In the last, some details and advantages of the proposed method are concluded.
\section{Preliminaries}
In this section, some concepts related to Intuitionistic Fuzzy Set (IFS) and Pythagorean Fuzzy Sets (PFS) are recalled.
\subsection{Intuitionistic Fuzzy Set}
The IFS $A$ defined in space X can be expressed as \cite{Atanassov1999}:
\begin{equation}
	A=\{{(x,\mu(x),\nu(x))|x\in X}\}
\end{equation}

$\mu(x)$ and $\nu(x)$ satisfy:
\begin{equation}
	\mu(x): X\to [0,1]  
\end{equation}
\begin{equation}
	\nu(x): X\to [0,1]  
\end{equation}

$\mu(x)$ is the degree of membership of $x \in X$ and  $\nu(x)$ is the degree of non-membership of $x \in X$. Both of them meet the condition that:
\begin{equation}
	0 \leq \mu(x) + \nu(x) \leq 1
\end{equation}

The hesitance function of an IFS $A$ in $X$ is defined as:
\begin{equation}
	\pi(x) = 1 - \mu(x) - \nu(x)
\end{equation}

The value of $\pi(x)$ can reflect the degree of hesitance of $x \in X$.
\subsection{Pythagorean fuzzy sets}
The PFS $A$ defined in space X can be expressed as \cite{Yager2014a}:
\begin{equation}
	A=\{{(x,A_{Y}(x),A_{N}(x)) | x\in X}\}
\end{equation}

Besides, $A_{Y}(x)$ and $A_{N}(x)$ satisfy:
\begin{equation}
	A_{Y}(x): X\to [0,1]  
\end{equation}
\begin{equation}
	A_{N}(x): X\to [0,1]  
\end{equation}

$A_{Y}(x)$ is a representative of the degree of membership of $x \in X$ and  $\nu(x)$ is a representative of the degree of non-membership of $x \in X$. Both of them meet the condition that:
\begin{equation}
	0 \leq A^{2}_{Y}(x) + A^{2}_{N}(x) \leq 1
\end{equation}

The hesitance function of an PFS $A$ in $X$ is defined as:
\begin{equation}
	A_{H}(x) = \sqrt{1 - A^{2}_{Y}(x) - A^{2}_{N}(x)}
\end{equation}

The value of $A_{H}(x)$ can reflect the degree of hesitance of $x \in X$.

\textbf{Property:}
Let $B$ and $C$ be two PFS, then both of them satisfy:
\begin{spacing}{1.7}
	\small \emph{(1) $B \subseteq C \ if \ \forall x \in X \ B_{Y}(x) \leq C_{Y}(x) \ and \ B_{N}(x) \geq C_{N}(x)$}
	
	\emph{(2) $B = C \ if \ \forall x \in X \ B_{Y}(x) = C_{Y}(x) \ and \ B_{N}(x) = C_{N}(x)$}
	
	\emph{(3) $B \ \cap \ C = \{\langle x, min[B_{Y}(x),C_{Y}(x)], max[B_{N}(x),C_{N}(x)] \rangle \\ | x \in X\}$}
	
	\emph{(4) $B \ \cup \ C = \{\langle x, max[B_{Y}(x),C_{Y}(x)], min[B_{N}(x),C_{N}(x)] \rangle \\ | x \in X\}$}
	
	\emph{(5)$C = \{\langle x, B_{Y}(x)C_{Y}(x), (B^{2}_{Y}(x) + C^{2}_{Y}(x) - B^{2}_{Y}(x)C^{2}_{Y}(x))^{\frac{1}{2}}\rangle  | x \in X\}$}
	
	\emph{(6) $B^{n} = \{\langle x, B^{n}_{Y}(x), (1 - (1 - B^{2}_{N}(x)^{n}))^{\frac{1}{2}}\rangle  | x \in X\}$}
\end{spacing}

In the sequel, some distances between different IFSs and PFSs are briefly introduced. Related function is also mentioned, which is helpful in judging the validity of IFS and PFS in its extended form.

\subsection{Distances Related to Pythagorean Fuzzy Sets}
The Hamming distance which measures difference between IFSs $D$ and $E$ is \cite{SZMIDT2000505}:
\begin{equation}
	\begin{split}
		d_{Hm}(D,E) = \frac{1}{2} \cdot(|\mu_{D}(x) - \mu_{E}(x)| + \\|\nu_{D}(x) - \nu_{E}(x)| +  |\pi_{D}(x) - \pi_{E}(x)|)
	\end{split}
\end{equation}

The normalized Hamming distance is expressed as \cite{SZMIDT2000505}:
\begin{equation}
	\begin{split}
		\widetilde{d}_{Hm}(D,E) = \frac{1}{2n} \cdot \sum_{i = 1}^{n}(|\mu_{D}(x_{i}) - \mu_{E}(x_{i})| + \\|\nu_{D}(x_{i}) - \nu_{E}(x_{i})| +  |\pi_{D}(x_{i}) - \pi_{E}(x_{i})|)
	\end{split}
\end{equation}

The Euclidean distance which measures difference between IFSs $D$ and $E$ is \cite{SZMIDT2000505}:
\begin{equation}
	\begin{aligned}
		d_{Eu}(D,E) = (\frac{1}{2} \cdot ((\mu_{D}(x) - \mu_{E}(x))^{2} + \\ (\nu_{D}(x) - \nu_{E}(x))^{2} + (\pi_{D}(x) - \pi_{E}(x))^{2}))^{\frac{1}{2}}
	\end{aligned}
\end{equation}

The normalized Euclidean distance is expressed as \cite{SZMIDT2000505}:
\begin{equation}
	\begin{aligned}
		\widetilde{d}_{Eu}(D,E) = (\frac{1}{2n} \cdot \sum_{i = 1}^{n}((\mu_{D}(x_{i}) - \mu_{E}(x_{i}))^{2} + \\ (\nu_{D}(x_{i}) - \nu_{E}(x_{i}))^{2} + (\pi_{D}(x_{i}) - \pi_{E}(x_{i}))^{2}))^{\frac{1}{2}}
	\end{aligned}
\end{equation}
\subsection{Extended Distance Measure of PFSs}
The Hamming distance which measures difference between PFSs $F$ and $G$ is \cite{CHEN2018129}:
\begin{equation}
	\begin{split}
		D_{Hm}(F,G) = \frac{1}{2} \cdot(|F^{2}_{Y}(x) - G^{2}_{Y}(x)| + \\|F^{2}_{N}(x) - G^{2}_{N}(x)| +  |F^{2}_{H}(x) - G^{2}_{H}(x)|)
	\end{split}
\end{equation}

The normalized Hamming distance is expressed as:
\begin{equation}
	\begin{split}
		\widetilde{D}_{Hm}(F,G) = \frac{1}{2n} \cdot \sum_{i = 1}^{n}(|F^{2}_{Y}(x) - G^{2}_{Y}(x)| + \\|F^{2}_{N}(x) - G^{2}_{N}(x)| +  |F^{2}_{H}(x) - G^{2}_{H}(x)|)
	\end{split}
\end{equation}

The Euclidean distance which measures difference between PFSs $F$ and $G$ is \cite{CHEN2018129}:
\begin{equation}
	\begin{aligned}
		D_{Eu}(F,G) = (\frac{1}{2} \cdot ((F^{2}_{Y}(x) - G^{2}_{Y}(x))^{2} + \\ (F^{2}_{N}(x) - G^{2}_{N}(x))^{2} + (F^{2}_{H}(x) - G^{2}_{H}(x))^2)^{\frac{1}{2}}
	\end{aligned}
\end{equation}

The normalized Euclidean distance is expressed as:
\begin{equation}
	\begin{aligned}
		\widetilde{D}_{Eu}(F,G) = (\frac{1}{2n} \cdot \sum_{i = 1}^{n}((F^{2}_{Y}(x) - G^{2}_{Y}(x))^{2} + \\ (F^{2}_{N}(x) - G^{2}_{N}(x))^{2} + (F^{2}_{H}(x) - G^{2}_{H}(x))^{2})^{\frac{1}{2}}
	\end{aligned}
\end{equation}

The Chen's distance which measures difference between PFSs $F$ and $G$ is \cite{CHEN2018129}:
\begin{equation}
	\begin{aligned}
		D_{C}(F,G) = [\frac{1}{2} \cdot (|F^{2}_{Y}(x) - G^{2}_{Y}(x)|^{\beta} + \\ |F^{2}_{N}(x) - G^{2}_{N}(x)|^{\beta} + |F^{2}_{H}(x) - G^{2}_{H}(x)|^{\beta})]^{\frac{1}{\beta}}
	\end{aligned}
\end{equation}

The normalized Chen's distance is expressed as:
\begin{equation}
	\begin{aligned}
		\widetilde{D}_{C}(F,G) = [\frac{1}{2n} \cdot \sum_{i = 1}^{n}(|F^{2}_{Y}(x) - G^{2}_{Y}(x)|^{\beta} + \\ |F^{2}_{N}(x) - G^{2}_{N}(x)|^{\beta} + |F^{2}_{H}(x) - G^{2}_{H}(x)|^{\beta})]^{\frac{1}{\beta}}
	\end{aligned}
\end{equation}

The parameter $\beta$ in Chen's distance which satisfies the condition that $\beta \geq 1$. When $\beta$ = 1, the method of Chen reduces to the method of Hamming distance. Besides, when  $\beta$ = 2, the method of Chen reduces to the method of Euclidean distance.

\subsection{Score Function on IFS}
A score function of IFS is initialized by Chen and Tan and it is utilized in the solution of problems which are produced in multi-attribute decision using intuitionistic sets. Let $A$ be an intuitionistc fuzzy set on the universe X $=\{x_{1},x_{2},...x_{n}\}$. IFS $A$ is given as $A = \{\langle x,\mu(x),\nu(x)\}$. The score function $S_{A}$ is defined as \cite{article5}:
\begin{equation}
	S_{A}(x_{i})=\mu(x_{i})-\nu(x_{i})
\end{equation}

The value of the formula illustrates a degree that whether the intuitionistic fuzzy sets is comfortable enough for decision makers to have a clear and straightforward expectation to actual situations. Obviously, the value of $\mu(x)$ gets larger, the smaller the value of $\nu(x)$ is going to be and the probability of $x \in X$ gets greater, which means the value of $S_{A}(x_{i})$ gets bigger. Therefore, the value of score function $S_{A}(x_{i})$ can be regarded as a kind of level of support about the element $x \in X$. When $S_{A}(x_{i}) > 0$, it can be concluded that it is more believable to classify $x$ as a component of $X$. When $S_{A}(x_{i}) < 0$, it can be concluded that it is more believable to classify $x$ as an exception of $X$. More than that, the absolute value of the score function can manifest the situation of the degree of certainty of IFS. If the mass is getting bigger, then the IFS is more certain. If the mass is getting smaller, then the IFS is more uncertain. For example, let $A$ and $B$ be two PFSs and they are defined respectively as $A = \{\langle x, 0.35, 0.35\rangle\}$ and $B = \{\langle x, 0.6, 0.1\rangle\}$. It is worth noting that the degree of hesitance in these two IFSs is exactly identical. However, the IFS $B$ is more valuable and useful, because it offers more information and can clearly illustrate actual situations. Obviously, the values of $|S_{A}(x) = 0|$ and $|S_{B}(x) = 0.5|$ is exactly consistent with the conclusion mentioned above. As a consequence, the absolute value of $S_{A}(x)$ can be utilized as an efficient method to measure the level of certainty.
\section{Proposed method}
How to measure the differences between PFSs is still an open issue. In this section, a proposed method of distance measure between PFSs. It aims to further improve the performance of the similarity matrix in managing the mass of membership, non-membership and the hesitance. Then, the properties of the proposed method are inferred and proven. In numerical examples, the proposed method designed for PFSs satisfies the distance measure axiom and is more efficient in producing intuitive and rational results.
\subsection{Extended Score Function for PFS}
On the base of the definition of score function which is developed on the concept of intuitionistic fuzzy set, a new score function $SP$ is on the notion of  pythagorean fuzzy set which is defined as:
\begin{equation}
	SP_{A}(x_{i})=A_{Y}^{2}(x_{i})-A_{N}^{2}(x_{i})
\end{equation}

The extended PFS formula plays a role similar to that of a scoring function based on IFS when dealing with PFSs and it is utilized to serve as a component of the proposed distance measure.
\subsection{The New Matrix-based Distance Measure}
Let $A$ and $B$ be two PFSs in the finite universe of discourse X, the new distance measure is defined as follows:
\begin{equation}
	D_{N}(A,B) = \sqrt{\frac{1}{n}\sum_{i=1}^{n}\frac{\vec{m}_{i}M(\vec{m}_{i}M)^{T}}{AB_Y+AB_N+AB_H}}
\end{equation}
\begin{equation}
	AB_Y = A^{4}_{Y}(x_{i})+B^{4}_{Y}(x_{i})
\end{equation}
\begin{equation}
	AB_N=	A^{4}_{N}(x_{i})+B^{4}_{N}(x_{i})
\end{equation}
\begin{equation}
	AB_H = A^{4}_{H}(x_{i})+B^{4}_{H}(x_{i})
\end{equation}
\begin{equation}\small
	\vec{m}_{i} = (A^{2}_{Y}(x_{i})-B^{2}_{Y}(x_{i}),A^{2}_{N}(x_{i})-B^{2}_{N}(x_{i}),A^{2}_{H}(x_{i})-B^{2}_{H}(x_{i}))
\end{equation}

However, the definition of the matrix $M$ is still not clear. The most important efficacy of the matrix is to adjust and optimize the mass of membership, non-membership and the hesitance. Adopting the index of hesitance as a parameter to generating distance between different PFSs is not straightforward and concise, because the index of hesitance itself is a kind of uncertainty, which is very difficult to clarify the relationship between different hesitance. There is a kind of relationship between the hesitance and membership and non-membership. For example, let $A$ and $B$ be two PFSs in the finite universe of discourse X, where PFS $A = \{\langle0,0,1\rangle\}$ and PFS $B = \{\langle0.51,0.49,0.706\rangle\}$. In PFS $A$, the index of hesitance is equal to 1, which indicates that PFS $A$ is totally uncertain. Besides, on the contrary, the index of hesitance of PFS $B$ is 0, it means there is no uncertainty or vagueness commonly. However, the mass of membership and non-membership is so close that it is very difficult to make reasonable decisions. It can be pointed out, the index of hesitance is not independent of the other two parameters, namely membership and non-membership, which means all of them have a similarity under certain relationship and is already sufficiently demonstrated in examples provided. Therefore, the definition of the new matrix $M$ is written as follows:
\begin{equation}
	M = \begin{pmatrix} 1 & 0 & 0 \\ 0 & 1 & 0 \\ Y & N & H \end{pmatrix}	
\end{equation}
\begin{equation}
	Y = \frac{(A^{2}_{Y}(x_{i})+B^{2}_{Y}(x_{i}))}{(A^{2}_{Y}(x_{i})+B^{2}_{Y}(x_{i}) + A^{2}_{N}(x_{i})+B^{2}_{N}(x_{i}))}
\end{equation}
\begin{equation}
	N = \frac{(A^{2}_{N}(x_{i})+B^{2}_{N}(x_{i}))}{(A^{2}_{Y}(x_{i})+B^{2}_{Y}(x_{i}) + A^{2}_{N}(x_{i})+B^{2}_{N}(x_{i}))}
\end{equation}
\begin{equation}
	H = \sqrt{1-Y^{2}-N^{2}}
\end{equation}

The main idea of the definition is that because uncertainty can not accurately show the real situation, then the mass of index is supposed to be distributed to membership and non-membership according to their mass in different PFSs to strengthen their ability of identification and not to change the original conditions, which enlarges the useful amount of information of pythagorean fuzzy sets and is helpful in target recognition. Due to the distribution to membership and non-membership, an adjustment of the index of hesitance should also be considered. Therefore, considering the mathematic form of PFS, it is proposed that $H = \sqrt{1-Y^{2}-N^{2}}$ is adopted as the remaining amount of information after the distribution to membership and non-membership. The operation of reduction in the index of hesitance improves the degree of identification, which is very helpful in measuring the distance of PFSs.

\textbf{Specific details about matrix: }The new distance proposed in this paper is based on the transformation of vectors from 3 parameters in PFSs. However, the role of the vector and matrix is not just showing the original mass of 3 parameters. What can be concluded is that membership and non-membership is independent of each other, so both of the parameters is considered as orthogonal. Though membership and non-membership are treated equally, the relationship between hesitance and the other 2 parameters is not just orthogonal. It presents a kind of proportion in membership and non-membership and also conforms to the operation of some methods which handles the uncertainties of multiple elements propositions in evidence theory and is closely related to fuzzy sets. For example, two PFSs $A = \{\langle0.6,0.6,0.529\rangle\}$ and PFS $B = \{\langle0.3,0.3,0.905\rangle\}$ are given to explain the process of the proposed vector and the new matrix which adjust the distribution of 3 parameters. The process is presented as follows:
\begin{spacing}{1.5}
	$\!$$\!$$\!$$\!$$\!$$\!$$\!$$\!$$\!$	
	$\vec{m}_{i} = (A^{2}_{Y}(x_{i})-B^{2}_{Y}(x_{i}),A^{2}_{N}(x_{i})-B^{2}_{N}(x_{i}),A^{2}_{H}(x_{i})-B^{2}_{H}(x_{i})) = (0.27, 0.27, -0.54)$\\
	$Y = \frac{(A^{2}_{Y}(x_{i})+B^{2}_{Y}(x_{i}))}{(A^{2}_{Y}(x_{i})+B^{2}_{Y}(x_{i}) + A^{2}_{N}(x_{i})+B^{2}_{N}(x_{i}))} = \frac{0.6^{2}+0.3^{2}}{2(0.6^{2}+0.3^{2})} = 0.5$ \\
	$N = \frac{(A^{2}_{N}(x_{i})+B^{2}_{N}(x_{i}))}{(A^{2}_{Y}(x_{i})+B^{2}_{Y}(x_{i}) + A^{2}_{N}(x_{i})+B^{2}_{N}(x_{i}))} = \frac{0.6^{2}+0.3^{2}}{2(0.6^{2}+0.3^{2})} = 0.5$\\ 
	$H = \sqrt{1-Y^{2}-N^{2}} = 0.707$
\end{spacing}
\begin{spacing}{1.5}
	$\!$$\!$$\!$$\!$$\!$$\!$$\!$$\!$	\footnotesize{$\vec{m}_{i}\times\begin{pmatrix} 1 & 0 & 0 \\ 0 & 1 & 0 \\ Y & N & H \end{pmatrix}=(0.27, 0.27, -0.54) \times \begin{pmatrix} 1 & 0 & 0 \\ 0 & 1 & 0 \\ 0.5 & 0.5 & 0.707 \end{pmatrix} \\= (0,0,0.3818)$}
\end{spacing}
Then the construction of a crucial part of the numerator is completed.
\subsection{Demonstration of Proposed Method Properties}
In this part, many properties of the proposed method are verified, like symmetry and triangle inequality.

\textbf{Proof 1:} The commutative property is verified in this proof. Let $A$ and $B$ be two PFSs in the finite universe of discourse X. Considering $D_{N}(A,B)$ and $D_{N}(B,A)$, two equations can be obtained:
\begin{spacing}{1.5}
	$\!$$\!$$\!$$\!$$\!$$\!$$D_{N}(A,B) = \sqrt{\frac{1}{n}\sum_{i=1}^{n}\limits\frac{\vec{m}_{i}M(\vec{m}_{i}M)^{T}}{AB_Y+AB_N+AB_H}}$\\
	$ AB_Y = A^{4}_{Y}(x_{i})+B^{4}_{Y}(x_{i})$\\
	$AB_N=	A^{4}_{N}(x_{i})+B^{4}_{N}(x_{i})$\\
	$AB_H = A^{4}_{H}(x_{i})+B^{4}_{H}(x_{i})$\\
	$\vec{m}_{i} = (A^{2}_{Y}(x_{i})-B^{2}_{Y}(x_{i}),A^{2}_{N}(x_{i})-B^{2}_{N}(x_{i}),A^{2}_{H}(x_{i})-B^{2}_{H}(x_{i}))$\\
	$M = \begin{pmatrix} 1 & 0 & 0 \\ 0 & 1 & 0 \\ Y & N & H \end{pmatrix}$\\
	$	Y = \frac{(A^{2}_{Y}(x_{i})+B^{2}_{Y}(x_{i}))}{(A^{2}_{Y}(x_{i})+B^{2}_{Y}(x_{i}) + A^{2}_{N}(x_{i})+B^{2}_{N}(x_{i}))}$\\
	$N = \frac{(A^{2}_{N}(x_{i})+B^{2}_{N}(x_{i}))}{(A^{2}_{Y}(x_{i})+B^{2}_{Y}(x_{i}) + A^{2}_{N}(x_{i})+B^{2}_{N}(x_{i}))}$
\end{spacing}

\begin{spacing}{1.5}
	$\!$$\!$$\!$$\!$$\!$$\!$$D_{N}(B,A) = \sqrt{\frac{1}{n}\sum_{i=1}^{n}\limits\frac{\vec{m}_{i}M(\vec{m}_{i}M)^{T}}{BA_Y+BA_N+BA_H}}$\\
	$ BA_Y = B^{4}_{Y}(x_{i})+A^{4}_{Y}(x_{i})$\\
	$BA_N=	B^{4}_{N}(x_{i})+A^{4}_{N}(x_{i})$\\
	$BA_H = B^{4}_{H}(x_{i})+A^{4}_{H}(x_{i})$\\
	$\vec{m}_{i} = (B^{2}_{Y}(x_{i})-A^{2}_{Y}(x_{i}),B^{2}_{N}(x_{i})-A^{2}_{N}(x_{i}),B^{2}_{H}(x_{i})-A^{2}_{H}(x_{i}))$\\
	$M = \begin{pmatrix} 1 & 0 & 0 \\ 0 & 1 & 0 \\ Y & N & H \end{pmatrix}$\\
	$	Y = \frac{B^{2}_{Y}(x_{i})+A^{2}_{Y}(x_{i})}{(B^{2}_{Y}(x_{i})+A^{2}_{Y}(x_{i}) + B^{2}_{N}(x_{i})+A^{2}_{N}(x_{i}))}$\\
	$N = \frac{B^{2}_{N}(x_{i})+A^{2}_{N}(x_{i})}{(B^{2}_{Y}(x_{i})+A^{2}_{Y}(x_{i}) + B^{2}_{N}(x_{i})+A^{2}_{N}(x_{i}))}$
\end{spacing}

It can be easily concluded that, when PFS $A$ and $B$ interchange their places, the attributes $ AB_Y$, $AB_N$, $AB_H$, matrix $M$, $Y$ and $N$ do not change their values. But how the numerator $\vec{m}_{i}M(\vec{m}_{i}M)^{T}$ change is not very clear. Nevertheless, it can be also easily proved like this:
\begin{spacing}{1.5}
	$\!$$\!$$\!$$\!$$\!$$\!$$\!$$\!$	$\vec{m}_{i}M(\vec{m}_{i}M)^{T}$\\
	$= (A^{2}_{Y}(x_{i})-B^{2}_{Y}(x_{i}) + Y (A^{2}_{H}(x_{i})-B^{2}_{H}(x_{i})))^{2}\\+ (A^{2}_{N}(x_{i})-B^{2}_{N}(x_{i}) + N (A^{2}_{H}(x_{i})-B^{2}_{H}(x_{i})))^{2}\\+ ((B^{2}_{H}(x_{i})-A^{2}_{H}(x_{i}))(\sqrt{1-Y^{2}-N^{2}}))^{2}\\
	=(B^{2}_{Y}(x_{i})-A^{2}_{Y}(x_{i}) + Y (B^{2}_{H}(x_{i})-A^{2}_{H}(x_{i})))^{2}\\+ (B^{2}_{N}(x_{i})-A^{2}_{N}(x_{i}) + N (B^{2}_{H}(x_{i})-A^{2}_{H}(x_{i})))^{2}\\+ ((A^{2}_{H}(x_{i})-B^{2}_{H}(x_{i}))(\sqrt{1-Y^{2}-N^{2}}))^{2}$
\end{spacing}

Therefore, it is proved that the proposed method of distance measure between PFSs satisfy the property that $D_{N}(A,B)=D_{N}(B,A)$.

\textbf{Proof 2:} A simple example is offered to verify a basic attribute that is when 2 PFSs are exactly the same, their distance is coincidently 0. Let $A$ and $B$ become two identical PFSs, namely $\{\langle0.5,0.5,0.707\rangle\}$, their distance is supposed to be 0 intuitively. According to the definition of the vector $\vec{m}_{i}$, every component in the vector is equal to zero, which means the numerator $\vec{m}_{i}M(\vec{m}_{i}M)^{T}$ of the method of distance measure is zero and the value of the whole formula is also zero. In a wider range, when two PFS are perfectly consistent, $A^{2}_{Y}(x_{i})=B^{2}_{Y}(x_{i}),A^{2}_{N}(x_{i})=B^{2}_{N}(x_{i}),A^{2}_{H}(x_{i})=B^{2}_{H}(x_{i})$, it can be concluded that the value of $\vec{m}_{i}$ should be exactly zero. Therefore,the distance of the proposed method satisfies the property that $D_{N}(A,B) = 0$ if and only if $A=B$.

\textbf{Proof 3: } In this proof, the distance between 2 PFSs is demonstrated that $0 \leq D_{N}(A,B) \leq 1$. Let $A$ and $B$ be two PFSs in the finite universe of discourse X, then it can be obtained that:
\begin{spacing}{1.5}
	$\!$$\!$$\!$$\!$$\!$$\!$$\!$$\!$	$\vec{m}_{i}M \\= (A^{2}_{Y}(x_{i})-B^{2}_{Y}(x_{i}) + Y (A^{2}_{H}(x_{i})-B^{2}_{H}(x_{i})),\\A^{2}_{N}(x_{i})-B^{2}_{N}(x_{i}) + N (A^{2}_{H}(x_{i})-B^{2}_{H}(x_{i})),\\(B^{2}_{H}(x_{i})-A^{2}_{H}(x_{i}))(\sqrt{1-Y^{2}-N^{2}})$\\
	$\vec{m}_{i}M(\vec{m}_{i}M)^{T}\\=(A^{2}_{Y}(x_{i})-B^{2}_{Y}(x_{i}) + Y (A^{2}_{H}(x_{i})-B^{2}_{H}(x_{i})))^{2}\\+ (A^{2}_{N}(x_{i})-B^{2}_{N}(x_{i}) + N (A^{2}_{H}(x_{i})-B^{2}_{H}(x_{i})))^{2}\\+ ((B^{2}_{H}(x_{i})-A^{2}_{H}(x_{i}))(\sqrt{1-Y^{2}-N^{2}}))^{2}\\= (A^{2}_{Y}(x_{i})-B^{2}_{Y}(x_{i}))^{2}+(A^{2}_{N}(x_{i})-B^{2}_{N}(x_{i}))^{2}\\
	+ (A^{2}_{H}(x_{i})-B^{2}_{H}(x_{i}))^{2}\\+2Y(A^{2}_{Y}(x_{i})-B^{2}_{Y}(x_{i}))(A^{2}_{H}(x_{i})-B^{2}_{H}(x_{i}))\\+2N(A^{2}_{N}(x_{i})-B^{2}_{N}(x_{i}))(A^{2}_{H}(x_{i})-B^{2}_{H}(x_{i}))\\=A^{4}_{Y}(x_{i})+B^{4}_{Y}(x_{i})+A^{4}_{N}(x_{i})+B^{4}_{N}(x_{i})+A^{4}_{H}(x_{i})+B^{4}_{H}(x_{i})\\-2A^{2}_{Y}(x_{i})B^{2}_{Y}(x_{i})-2A^{2}_{N}(x_{i})B^{2}_{N}(x_{i})-2A^{2}_{H}(x_{i})B^{2}_{H}(x_{i})\\+(A^{2}_{H}(x_{i})-B^{2}_{H}(x_{i}))\\((2YA^{2}_{Y}(x_{i})+2NA^{2}_{N}(x_{i}))-(2YB^{2}_{Y}(x_{i})+2NB^{2}_{N}(x_{i}))) \leq A^{4}_{Y}(x_{i})+B^{4}_{Y}(x_{i})+A^{4}_{N}(x_{i})+B^{4}_{N}(x_{i})+A^{4}_{H}(x_{i})+B^{4}_{H}(x_{i})$
\end{spacing}

Obviously, when the entirety of $2YA^{2}_{Y}(x_{i})+2NA^{2}_{N}(x_{i})$ is larger than that of $2YB^{2}_{Y}(x_{i})+2NB^{2}_{N}(x_{i})$, $A^{2}_{H}(x_{i})-B^{2}_{H}(x_{i})$ is definitely negative, according to the definition of pythagorean fuzzy set. Therefore, it can be inferred that
\begin{spacing}{1.5}\small
	$0 \leq \frac{\vec{m}_{i}M(\vec{m}_{i}M)^{T}}{A^{4}_{Y}(x_{i})+B^{4}_{Y}(x_{i})+A^{4}_{N}(x_{i})+B^{4}_{N}(x_{i})+A^{4}_{H}(x_{i})+B^{4}_{H}(x_{i})}\leq 1$
	
	$0 \leq \frac{1}{n}\sum_{i=1}^{n} \limits\frac{\vec{m}_{i}M(\vec{m}_{i}M)^{T}}{A^{4}_{Y}(x_{i})+B^{4}_{Y}(x_{i})+A^{4}_{N}(x_{i})+B^{4}_{N}(x_{i})+A^{4}_{H}(x_{i})+B^{4}_{H}(x_{i})}\leq 1$
	
	Let $T = \frac{1}{n}\sum_{i=1}^{n} \limits\frac{\vec{m}_{i}M(\vec{m}_{i}M)^{T}}{A^{4}_{Y}(x_{i})+B^{4}_{Y}(x_{i})+A^{4}_{N}(x_{i})+B^{4}_{N}(x_{i})+A^{4}_{H}(x_{i})+B^{4}_{H}(x_{i})}$
	
	Therefore, $0 \leq \sqrt{T} \leq 1$
\end{spacing}

Because $\sqrt{T}$ has the same significance with $D_{n}(A,B)$. Then, it can be concluded that $0 \leq D_{n}(A,B) \leq 1$.

\textbf{Proof 4:} The triangle inequality is going to be proven in this part. Some assumptions are given as: 
\begin{spacing}{1.5}
	$\begin{aligned}
		&\qquad Assumption\ 1 : A_{Y}^{2}(x) \leq  B_{Y}^{2}(x) \leq  C_{Y}^{2}(x)\\
		&\qquad Assumption\ 2 : C_{Y}^{2}(x) \leq  B_{Y}^{2}(x) \leq  A_{Y}^{2}(x)\\
		&\qquad Assumption\ 3 : B_{Y}^{2}(x) \leq min\{A_{Y}^{2}(x),C_{Y}^{2}(x)\}\\
		&\qquad Assumption\ 4 : B_{Y}^{2}(x) \geq max\{A_{Y}^{2}(x),C_{Y}^{2}(x)\}
	\end{aligned}$
\end{spacing}

It can be easily demonstrated that the inequality:
\begin{spacing}{1.5}
	$\!$$\!$$\!$$\!$$\!$$\!$$|A_{Y}^{2}(x)- C_{Y}^{2}(x)| \leq |A_{Y}^{2}(x)- B_{Y}^{2}(x)| + |B_{Y}^{2}(x)- C_{Y}^{2}(x)|$
\end{spacing}
which meets conditions of $Assumption\ 1$ and $Assumption\ 2$. On the base of $Assumption\ 3$, it can be concluded that
\begin{spacing}{1.5}
	\qquad \qquad $A_{Y}^{2}(x) \geq B_{Y}^{2}(x)$, $C_{Y}^{2}(x) \geq B_{Y}^{2}(x)$
\end{spacing}

Therefore, it can be obtained that:
\begin{spacing}{1.5}
	$\!$$\!$$\!$$\!$$\!$$\!$$|A_{Y}^{2}(x) - B_{Y}^{2}(x)|+|B_{Y}^{2}(x) - C_{Y}^{2}(x)|-|A_{Y}^{2}(x) - C_{Y}^{2}(x)|\\=
	f(x)=\left\{
	\begin{aligned}
		A_{Y}^{2}(x)- B_{Y}^{2}(x) + C_{Y}^{2}(x)- B_{Y}^{2}(x)- A_{Y}^{2}(x)\\+ C_{Y}^{2}(x)\ \ if \  A_{Y}^{2}(x)\geq C_{Y}^{2}(x)\\
		A_{Y}^{2}(x)- B_{Y}^{2}(x) + C_{Y}^{2}(x)- B_{Y}^{2}(x)+A_{Y}^{2}(x)\\- C_{Y}^{2}(x)\ \ if \  A_{Y}^{2}(x)\leq C_{Y}^{2}(x)\\
	\end{aligned}
	\right.$\\\\
	$=2 \cdot (min\{A_{Y}^{2}(x), C_{Y}^{2}(x)\}-B_{Y}^{2}(x)) \geq 0$
\end{spacing}

Similarly, on the base of $Assumption\ 4$, it can be summarized that:
\begin{spacing}{1.5}
	$\!$$\!$$\!$$\!$$\!$$\!$$|A_{Y}^{2}(x) - B_{Y}^{2}(x)|+|B_{Y}^{2}(x) - C_{Y}^{2}(x)|-|A_{Y}^{2}(x) - C_{Y}^{2}(x)|\\=
	f(x)=\left\{
	\begin{aligned}
		B_{Y}^{2}(x)- A_{Y}^{2}(x) + B_{Y}^{2}(x)- C_{Y}^{2}(x)- A_{Y}^{2}(x)\\
		+ C_{Y}^{2}(x)\ \ if \  A_{Y}^{2}(x)\geq C_{Y}^{2}(x)\\
		B_{Y}^{2}(x)- A_{Y}^{2}(x) + B_{Y}^{2}(x)- C_{Y}^{2}(x)+A_{Y}^{2}(x)\\- C_{Y}^{2}(x)\ \ if \  A_{Y}^{2}(x)\leq C_{Y}^{2}(x)\\
	\end{aligned}
	\right.$\\\\
	$=2 \cdot (B_{Y}^{2}(x)-max\{A_{Y}^{2}(x), C_{Y}^{2}(x)\}) \geq 0$
\end{spacing}

Hence, the property of inequality is also satisfied under the condition provided by $Assumption\ 3$ and $Assumption\ 4$.
\begin{spacing}{1.5}
	$\!$$\!$$\!$$\!$$\!$$\!$$|A_{Y}^{2}(x)- C_{Y}^{2}(x)| \leq |A_{Y}^{2}(x)- B_{Y}^{2}(x)| + |B_{Y}^{2}(x)- C_{Y}^{2}(x)|$
\end{spacing}

Because every component in the method of measuring distance is correspondingly proportionable to membership, non-membership and the index of hesitance. As a result, the triangle inequality has been proved. It can be concluded that:
\begin{spacing}{1.5}
	\qquad \qquad $D_{N}(A,B)+D_{N}(B,C) \geq D_{N}(A,C)$
\end{spacing}

All of the required properties of the proposed which are expected to be satisfied have been demonstrated. In the next part, some examples are offered to further verify the validity of proposed method.

\section{Numerical examples and applications}
In this section, lots of numerical examples and actual applications are utilized to illustrate the effectiveness of the method proposed in this paper. And the accuracy and superiority of this method are also testified in this part.

\begin{table}[h]\small
	\centering
	\caption{Distances generated by Euclidean's method and proposed method}
	\begin{spacing}{1}
		\setlength{\tabcolsep}{1mm}{
			\begin{tabular}{c c c}\hline
				$The$ $change$ $of$ $\delta$ $value$&\multicolumn{2}{c}{$Distances$}\\\hline
				&$Euclidean$&$Proposed\ method$\\
				$0.1$&$0.1$&$0.1048$\\
				$0.2$&$0.1$&$0.1154$\\
				$0.3$&$0.1$&$0.1259$\\
				$0.4$&$0.1$&$0.1348$\\
				$0.5$&$0.1$&$0.1400$\\
				$0.6$&$0.1$&$0.1400$\\
				$0.7$&$0.1$&$0.1348$\\
				$0.8$&$0.1$&$0.1259$\\
				$0.9$&$0.1$&$0.1154$\\
				$1.0$&$0.1$&$0.1048$\\\hline
		\end{tabular}}
	\end{spacing}
	\label{table1}
\end{table}
\begin{table}[h]\small
	\centering
	\caption{The change of the value of score function}
	\begin{spacing}{1.20}
		\setlength{\tabcolsep}{0.8mm}{
			\begin{tabular}{c c c c c c c c c c c}\hline
				$\delta$&\multicolumn{1}{c}{$0.1$}&\multicolumn{1}{c}{$0.2$}&\multicolumn{1}{c}{$0.3$}&\multicolumn{1}{c}{$0.4$}&\multicolumn{1}{c}{$0.5$}&\multicolumn{1}{c}{$0.6$}&\multicolumn{1}{c}{$0.7$}&\multicolumn{1}{c}{$0.8$}&\multicolumn{1}{c}{$0.9$}&\multicolumn{1}{c}{$1$}\\\hline
				$S(A)$&$-0.8$&$-0.6$&$-0.4$&$-0.2$&$0$&$0.2$&$0.4$&$0.6$&$0.8$&$1$\\
				$S(B)$&$-1$&$-0.8$&$-0.6$&$-0.4$&$-0.2$&$0$&$0.2$&$0.4$&$0.6$&$0.8$\\
				$|S(A)|$&$0.8$&$0.6$&$0.4$&$0.2$&$0$&$0.2$&$0.4$&$0.6$&$0.8$&$1$\\
				$|S(B)|$&$1$&$0.8$&$0.6$&$0.4$&$0.2$&$0$&$0.2$&$0.4$&$0.6$&$0.8$\\\hline
		\end{tabular}}
	\end{spacing}
	\label{table2}
\end{table}
\begin{figure}
	\centering
	\includegraphics[scale=0.5]{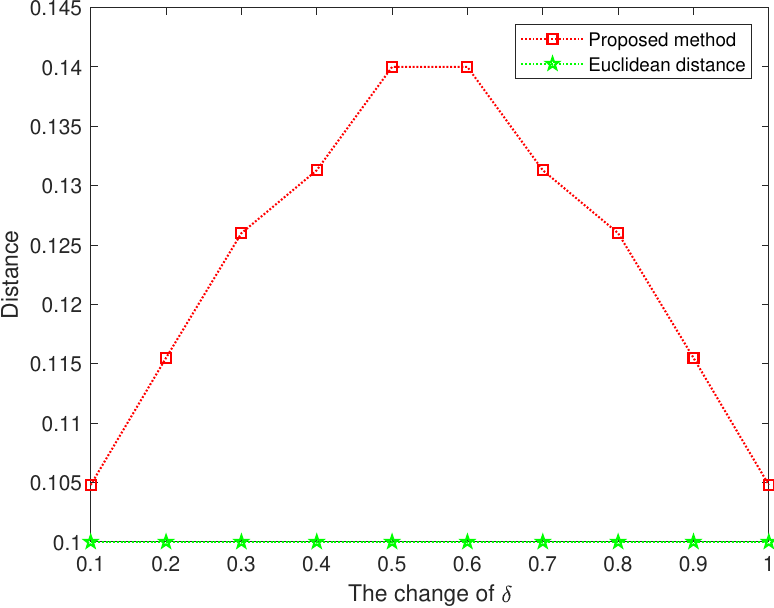}
	\caption{The visualized results of distances with the change of $\delta$ value}
	\label{fig1}
\end{figure}
\begin{figure}
	\centering
	\includegraphics[scale=0.5]{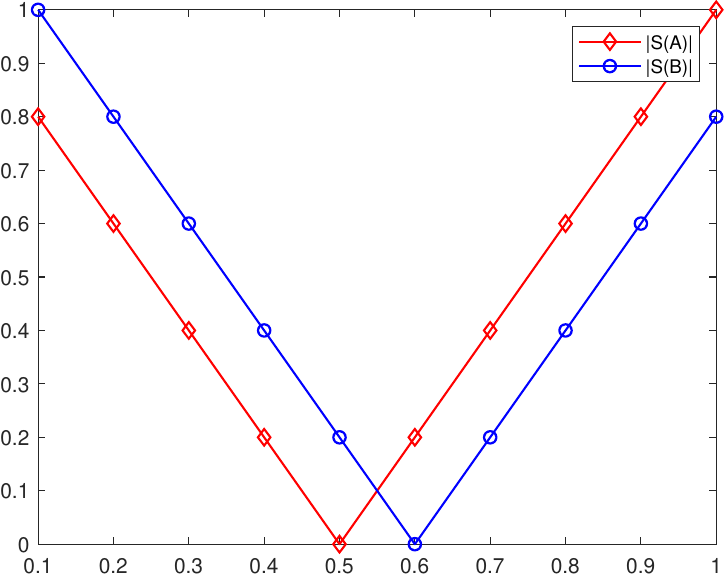}
	\caption{The visualized results of the change of the value of score function}
	\label{fig2}
\end{figure}
\begin{table*}[t]
	\caption{Examples of PFSs}
	\centering
	\begin{spacing}{1.20}
		\setlength{\tabcolsep}{8mm}{
			\begin{tabular}{lccccccccc}
				\midrule
				$PFSs$ & $Case 1$ & $Case 2$ \\
				\midrule
				$A_{i}$ & $\{\langle {\color{red} x_{1},0.55, 0.45}\rangle,\langle {\color{red}x_{2},0.63, 0.55}\rangle\}$ & $\{\langle {\color{red}x_{1},0.55, 0.45}\rangle,\langle {\color{red}x_{2},0.63, 0.55}\rangle\}$ \\
				$B_{i}$ & $\{\langle x_{1},0.39, 0.50\rangle,\langle x_{2},0.50, 0.59\rangle\}$ & $\{\langle x_{1},0.40, 0.51\rangle,\langle x_{2},0.51, 0.60\rangle\}$ \\
				\midrule
				$PFSs$ & $Case 3$ & $Case 4$ \\
				\midrule
				$A_{i}$ & $\{\langle {\color{blue}x_{1},0.71, 0.63}\rangle,\langle {\color{blue}x_{2},0.63, 0.55}\rangle\}$ & $\{\langle {\color{blue}x_{1},0.71, 0.63}\rangle,\langle {\color{blue}x_{2},0.63, 0.55}\rangle\}$ \\
				$B_{i}$ & $\{\langle x_{1},0.63, 0.63\rangle,\langle x_{2},0.71, 0.63\rangle\}$ & $\{\langle x_{1},0.77, 0.55\rangle,\langle x_{2},0.55, 0.45\rangle\}$ \\
				\midrule
				$	PFSs$ & $Case 5$ & $Case 6$ \\
				\midrule
				$A_{i}$ & $\{\langle {\color{red}x_{1},0.55, 0.45}\rangle,\langle {\color{red}x_{2},0.63, 0.55}\rangle\}$ & $\{\langle {\color{green}x_{1},0.30, 0.20}\rangle,\langle {\color{green}x_{2},0.40, 0.30}\rangle\}$ \\
				$B_{i}$ & $\{\langle x_{1},0.67, 0.39\rangle,\langle x_{2},0.74, 0.50\rangle\}$ & $\{\langle x_{1},0.15, 0.25\rangle,\langle x_{2},0.25, 0.35\rangle\}$ \\
				\midrule
				$PFSs$ & $Case 7$ & $Case 8$ \\
				\midrule
				$A_{i}$ & $\{\langle {\color{green}x_{1},0.30, 0.20}\rangle,\langle {\color{green}x_{2},0.40, 0.30}\rangle\}$ & $\{\langle x_{1},0.50, 0.40\rangle,\langle x_{2},0.40, 0.30\rangle\}$ \\
				$B_{i}$ & $\{\langle x_{1},0.45, 0.15\rangle,\langle x_{2},0.55, 0.25\rangle\}$ & $\{\langle x_{1},0.40, 0.40\rangle,\langle x_{2},0.50, 0.40\rangle\}$ \\
				\hline
		\end{tabular}}
	\end{spacing}
	\label{table3}
\end{table*}
\begin{table*}[t]
	\caption{Distances generated by different methods}
	\centering
	\begin{spacing}{1.20}
		\setlength{\tabcolsep}{3mm}{
			\begin{tabular}{lccccccccc}
				\midrule
				$Methods$ & $Case 1$ & $Case 2$ & $Case 3$ & $Case 4$ & $Case 5$ & $Case 6$ & $Case 7$ & $Case 8$\\
				\midrule
				$\widetilde{d}_{Hm}$ & $0.1464$ & $0.1351$ & $0.1170$ & $0.1350$ & $0.1161$ & {\color{red}$0.1500$} & {\color{red}$0.1500$} & {\color{red}$0.1500$} \\
				$\widetilde{d}_{Eu}$ & $0.1298$ & $0.1180$ & $0.1110$ & $0.1261$ & $0.1009$ & {\color{green}$0.1323$} & {\color{green}$0.1323$} & $0.1414$ \\
				$\widetilde{D}_{Hm}$ & {\color{red}$0.1500$} & $0.1400$ & {\color{red}$0.1500$} & {\color{red}$0.1500$} & {\color{red}$0.1500$} & $0.0825$ & $0.1275$ & $0.1250$ \\
				$\widetilde{D}_{Eu}$ & {\color{green}$0.1323$} & $0.1217$ & {\color{blue}$0.1414$} & {\color{blue}$0.1414$} & {\color{green}$0.1323$} & $0.0740$ & $0.1186$ & $0.1170$ \\
				$\widetilde{D}_{C}(\beta = 1)$ & {\color{red}$0.1500$} & $0.1400$ & {\color{red}$0.1500$} & {\color{red}$0.1500$} & {\color{red}$0.1500$} & $0.0825$ & $0.1275$ & $0.1250$ \\
				$\widetilde{D}_{C}(\beta = 2)$ & {\color{green}$0.1323$} & $0.1217$ & {\color{blue}$0.1414$} & {\color{blue}$0.1414$} & {\color{green}$0.1323$} & $0.0740$ & $0.1186$ & $0.1170$ \\
				$\widetilde{D}_{N}$ & $0.1801$ & $0.1756$ & $0.1417$ & $0.1517$ & $0.1578$ & $0.0746$ & $0.0829$ & $0.0982$ \\
				\hline
		\end{tabular}}
	\end{spacing}
		\label{table4}
\end{table*}
\subsection{Examples and Discussions}
\textbf{Example 1:} Let $A$ and $B$ be two PFSs in the finite universe of discourse X, which are defined as $A=\{(x,A_{Y}(x),A_{N}(x))\}$ and $B=\{(x,B_{Y}(x), B_{N}(x))\}$. Besides, a further limitation is defined as:
\begin{spacing}{1.5}
	$\begin{aligned}
		&\qquad A_{Y}^{2}(x)=\delta, A_{N}^{2}(x) = 1 - \delta, A_{Y}^{2}(x) = \delta - 0.1,\\
		&\qquad \ \ \ \ A_{N}^{2}(x) = 1.1 - \delta, A_{H}^{2}(x) = 0 = B_{H}^{2}(x)
	\end{aligned}$
\end{spacing}

The results of distances with the change of $\delta$ value and the change of the value of score function are provided in Fig \ref{fig1} and \ref{fig2}. When setting the index of hesitance to 0, the difference between 2 PFSs is only about the values of membership and non-membership. The operation is carried to simplify the process of calculation. Of course, any parameter can be modified in this way, as long as the mass of each parameter satisfies the properties of  pythagorean fuzzy set. The difference of classic Euclidean distance and the proposed is going to be discussed. All the results generated by the two methods are shown in Table \ref{table1}. The values obtained by score function are also shown in Table  \ref{table2} with the consistent increase of the value of parameter $\delta$ which indicates a corresponding change in the distance of IFSs.

It can be easily noted that the distance generated by classic Euclidean's method is a fixed value. However, this phenomenon is conflicting with the intuitive judgements. With the change of the $\delta$, if the results have not changed, then the results produced should be regarded irrational and counter-intuitive, which also indicates that the classic Euclidean distance is not sensitive to tiny variation among different PFSs. It may not accurately reveal potential changes in the relation of PFSs when they vary correspondingly but not symmetrically, which is crucial about whether a method of measuring distance can make reasonable judgements. Besides, when coming to analyze the variation of the variable $\delta$, the rationality and correctness of the proposed method is highlighted. With the increase of parameter $\delta$, when $\delta$ is in the range $0.1$ to $0.5$, the level of vagueness in these PFSs is getting higher, so the distance between two PFSs reaches its zenith. That is because the proposed method can better detect the changes of different PFSs and present the difference in the results produced, which improves the sensitivity of distance measure compared to classic Euclidean distance. The trend of the changes of PFSs can also be shown by the variation of the value of score function. While the parameter gradually reached the value of $0.5$, both of the absolute values of the score function of PFSs $A$ and $B$ are becoming smaller, which indicates the level of ambiguity of both PFSs is also getting higher. Because the PFSs are getting more uncertain, the distance between them is becoming bigger to indicate their distance can not be properly measured. Therefore, the proposed method reflects this factor in the results and can better embody underlying relationship between or among PFSs. 

\textbf{Example 2:} In this part, some numerical examples are given to illustrate the superiority of proposed method compared to other previous methods. Let $A_{i}$ and $B_{i}$ be two PFSs in the finite universe of discourse $X = \{x_{1},x_{2}\}$ and all of the docimastic cases are presented in Table \ref{table3}. And the distances generated by different methods are shown in Table \ref{table4}. When coming to analyze the examples given in Table \ref{table3}, it can be easily concluded that $A_{1} = A_{2} =A_{5}$, but $B_{1} \neq B_{2} \neq B_{5}$; $A_{3} = A_{4}$, but $B_{3} \neq B_{4}$; $A_{6} = A_{7}$, but $B_{6} \neq B_{7}$. However, the results produced by different methods show a completely different figure among provided methods of measuring distances. After inspecting all of results, it can be summarized that:

\textbf{1.} After comparing all of the method of measuring distances between different PFSs, every method can accurately detect the differences under the condition in $Case 1$ and $Case 2$.

\textbf{2.} However, the results produced by $\widetilde{D}_{Hm}$, $\widetilde{D}_{Eu}$, $\widetilde{D}_{C}(\beta = 1)$ and $\widetilde{D}_{C}(\beta = 2)$ seem not satisfying. They produce the exactly the same results in completely different cases, which is counter-intuitive and irrational.

\textbf{3.} Besides, when checking the results produced by $\widetilde{d}_{Hm}$ and $\widetilde{d}_{Eu}$ in $Case 6$ and $Case 7$, they are also unreasonable due to the same results in different cases.

\textbf{4.} More than what have been mentioned in point $3$, $\widetilde{d}_{Hm}$ also dose not perform well in $Case 7$ and $Case 8$. Because the method produces another counter-intuitive results under different conditions. 

\textbf{5.} All in all, when comparing all the cases mentioned above, it is very easy to distinguish that the distance of the proposed method is more sensitive to the changes in PFSs and significantly indicates the discrimination level of different PFSs, which is much more efficient than any other methods. It is more rational and closer to actual situations.

\textbf{6.} A more important point should be noticed is that the proposed method performs well under any cases  given in the table 3 in producing distances between PFSs, which other ones generate irrational and counter-intuitive results in the same cases. 
\begin{table}[h]\scriptsize
	\centering
	\caption{Properties satisfied by different methods}
	\setlength{\tabcolsep}{1mm}{
		\begin{spacing}{1.40}
			\begin{tabular}{c c c c c}\hline
				\multirow{3}*{$Methods$}&\multicolumn{4}{c}{$Properties$}\\\cline{2-5}
				&\makecell{$Non$$-$\\$degeneracy$}& $Symmetry$&\makecell{$Triangle$\\ $inequality$}& $Boundness$\\\hline
				$\widetilde{d}_{Hm}$&$\surd$&$\surd$&$\surd$&$\surd$\\
				$\widetilde{d}_{Eu}$&$\surd$&$\surd$&$\surd$&$\surd$\\
				$\widetilde{D}_{Hm}$&$\surd$&$\surd$&$\surd$&$\surd$\\
				$\widetilde{D}_{Eu}$&$\surd$&$\surd$&$\surd$&$\surd$\\
				$\widetilde{D}_{C}(\beta = 1)$&$\surd$&$\surd$&$\surd$&$\surd$\\
				$\widetilde{D}_{C}(\beta = 2)$&$\surd$&$\surd$&$\surd$&$\surd$\\
				$\widetilde{D}_{N}$&$\surd$&$\surd$&$\surd$&$\surd$\\\hline
	\end{tabular}\end{spacing}}
\label{table5}
\end{table}

\textbf{Notice: } It can be inferred from Table \ref{table5} that except for the normalized Chen's distance measure $\widetilde{D}_{C}$, other methods of measuring distances like $\widetilde{d}_{Hm}$, $\widetilde{d}_{Eu}$, $\widetilde{D}_{Hm}$, $\widetilde{D}_{Eu}$, $\widetilde{D}_{N}$ satisfy all the properties which the methods of distance measure are supposed to have. Moreover, the reason for the proposed method can generate more rational and intuitive results is that the proposed method considers discrepancies in PFSs and enlarges their influences in producing distances, which plays a significant role in manifesting differences. As a result, the proposed method is more acceptable and conforms to actual situations.

\textbf{Example 3:} Suppose there are three elements in the form of PFSs within the finite universe of discourse $A= \{a_1, a_2, a_3\}$. Then, we consider an application on the task of pattern recognition and the patterns $\mathcal{C} = \{\mathcal{C}_1, \mathcal{C}_2, \mathcal{C}_3, \mathcal{C}_4\}\}$ expressed by PFSs are present in Table \ref{fgfgfgf}. The aim is to classify a unrecognized pattern which is given as:
\begin{spacing}{1.5}
	$\mathcal{C}_{un1} = \{\langle x_{1},0.45, 0.39\rangle, \langle x_{2},0.67, 0.22\rangle, \langle x_{3},0.71, 0.23\rangle\}$
\end{spacing}
\begin{spacing}{1.1}
	$\mathcal{C}_{un2} = \{\langle x_{1},0.56, 0.26\rangle, \langle x_{2},0.29, 0.22\rangle, \langle x_{3},0.87, 0.14\rangle\}$
\end{spacing}
\begin{table}[H]
	\caption{Symptoms expressed in the form of PFS of patients in example 3}
	\centering
	\begin{spacing}{1.20}
		\setlength{\tabcolsep}{1.2mm}{
			\begin{tabular}{cccccccccc}
				\midrule
				$Patterns$ & $Attribute 1$ & $Attribute 2$ & $Attribute 3$  \\
				\midrule
				$\mathcal{C}_{1}$ & $\langle x_{1},0.70, 0.20\rangle$ & $\langle x_{2},0.65, 0.12\rangle$ & $\langle x_{3},0.28, 0.60\rangle$ \\
				$\mathcal{C}_{2}$ & $\langle x_{1},0.05, 0.70\rangle$ & $\langle x_{2},0.55, 0.74\rangle$ & $\langle x_{3},0.68, 0.12\rangle$ \\
				$\mathcal{C}_{3}$ & $\langle x_{1},0.82, 0.19\rangle$ & $\langle x_{2},0.74, 0.19\rangle$ & $\langle x_{3},0.24, 0.73\rangle$ \\
				$\mathcal{C}_{4}$ & $\langle x_{1},0.58, 0.11\rangle$ & $\langle x_{2},0.67, 0.30\rangle$ & $\langle x_{3},0.38, 0.43\rangle$ \\
				\hline
		\end{tabular}}
	\end{spacing}
	\label{fgfgfgf}
\end{table}

\begin{table}[h]
	\centering
	\caption{Distances generated by proposed method in example 3 for $\mathcal{C}_{un1}$}
	\begin{spacing}{1.20}
		\begin{tabular}{c c c c c c}\hline
			$Pattern$ &\multicolumn{3}{c}{$Distances$}&$Total$\\\hline
			&$x_{1}$& $x_{2}$&$x_{3}$&\\
			$\mathcal{C}_{1}$&$0.2049$&$0.0043$&$0.5322$&$0.4971$\\
			$\mathcal{C}_{2}$&$0.3004$&$0.4499$&$0.0067$&$0.5023$\\
			$\mathcal{C}_{3}$&$0.3813$&$0.0180$&$0.7042$&$0.5881$\\
			$\mathcal{C}_{4}$&$0.0707$&$0.0040$&$0.3112$&$0.4343$\\\hline
		\end{tabular}
	\end{spacing}
	\label{table7}
\end{table}

\begin{table}[h]
	\centering
	\caption{Distances generated by proposed method in example 3 for $\mathcal{C}_{un2}$}
	\begin{spacing}{1.20}
		\begin{tabular}{c c c c c c}\hline
			$Pattern$ &\multicolumn{3}{c}{$Distances$}&$Total$\\\hline
			&$x_{1}$& $x_{2}$&$x_{3}$& \\
			$\mathcal{C}_{1}$&$0.0691$&$0.1788$&$0.7478$&$0.5761$\\
			$\mathcal{C}_{2}$&$0.5080$&$0.5697$&$0.1033$&$0.6274$\\
			$\mathcal{C}_{3}$&$0.2041$&$0.3140$&$0.8518$&$0.6823$\\
			$\mathcal{C}_{4}$&$0.0072$&$0.2771$&$0.5409$&$0.4977$\\\hline
		\end{tabular}
	\end{spacing}
	\label{table8}
\end{table}

After calculation, we can distinguish the differences between unrecognized patterns and reference patterns and the results are provided in Table \ref{table7} and \ref{table8}. Through the analysis of tabular data, it is evident that the disparities between the first, second, and third patterns compared to the pattern to be identified are greater than those of the fourth pattern. In comparison to the reference pattern of the fourth type, the differences between the two types of unrecognized patterns are smaller. It can be concluded that the two patterns to be identified should belong to the reference fourth type.

\subsection{Algorithm Designed for Decision-making Problems}
\label{secal}
\textbf{Problem statement:} Assume there are a finite universe of discourse $X = \{x_{1},x_{2},x_{3},...,x_{n}\}$ and existing medical patterns $P = \{P_{1},P_{2},P_{3},...,P_{k}\}$ consisting of n elements in the form of PFSs, expressed as $P_{j} = \{\langle x_{i}, P_{Y}^{j}(x_{i}), P_{N}^{j}(x_{i})\} (1 \leq j \leq k)$ within the finite universe of discourse $X$. Several examples $E = \{E_{1}, E_{2}, E_{3},...,E_{r}\}$ which is composed of $r$ samples is given to be recognized and testify the correctness of the new algorithm. All of the elements in example $E$ is denoted as the form of PFS and the whole example is written as $E_{u} = \{\langle x_{i}, E_{Y}^{u}(x_{i}), E_{N}^{u}(x_{i})\} (1 \leq u \leq r)$. In sum, what are expected to be achieved is to decide or classify every element in example $E_{u}$ whether belongs to the pattern $P_{j}$. The algorithm is designed as:

\textbf{Step 1: } For every element in $E_{u}$, the proposed method of measuring distances is utilized to produce the distance between $P_{j}$ and $E_{u}$.
\begin{spacing}{1.5}\small
	$D_{N}(P_{j},E_{u}) = \sqrt{\frac{1}{n}\sum_{i=1}^{n}\frac{\vec{m}_{i}M(\vec{m}_{i}M)^{T}}{P_{j}E_{u}Y+P_{j}E_{u}N+P_{j}E_{u}H}}$
	
	$P_{j}E_{u}Y = P_{j}^{4}{Y}(x_{i})+E_{u}^{4}{Y}(x_{i})$
	
	$
	P_{j}E_{u}N=	P_{j}^{4}{N}(x_{i})+E_{u}^{4}{N}(x_{i})
	$
	
	$
	P_{j}E_{u}H = P_{j}^{4}{H}(x_{i})+E_{u}^{4}{H}(x_{i})
	$
	
	$
	\vec{m}_{i} = (P_{j}^{2}{Y}(x_{i})-E_{u}^{2}{Y}(x_{i}),P_{j}^{2}{N}(x_{i})-E_{u}^{2}{N}(x_{i}),P_{j}^{2}{H}(x_{i})-E_{u}^{2}{H}(x_{i}))
	$
	
	$
	M = \begin{pmatrix} 1 & 0 & 0 \\ 0 & 1 & 0 \\ Y & N & H \end{pmatrix}	
	$
	
	$
	Y = \frac{(P_{j}^{2}{Y}(x_{i})+E_{u}^{2}{Y}(x_{i}))}{(P_{j}^{2}{Y}(x_{i})+E_{u}^{2}{Y}(x_{i}) + P_{j}^{2}{N}(x_{i})+E_{u}^{2}{N}(x_{i}))}
	$
	
	$
	N = \frac{(P_{j}^{2}{N}(x_{i})+E_{u}^{2}{N}(x_{i}))}{(P_{j}^{2}{Y}(x_{i})+E_{u}^{2}{Y}(x_{i}) + P_{j}^{2}{N}(x_{i})+E_{u}^{2}{N}(x_{i}))}
	$
	
	$
	H = \sqrt{1-Y^{2}-N^{2}}
	$
\end{spacing}
\textbf{Step 2 : } After calculating the distance of every pair of $P_{j}$ and $E_{u}$, the smallest value of the distances between two PFSs $P_{j}$ and $E_{u}$ is selected, which is written as:
\begin{spacing}{1.5}
	\qquad\qquad\quad	$\widetilde{D}_{N}^{chosen} = \mathop{min}\limits_{1\leq u\leq r}\widetilde{D}_{N}(P_{j},E_{u})$
\end{spacing}

\textbf{Step 3 : } According to the results generated by \textbf{Step 2}, an element is classified into a pattern $P_{\beta}$, which is written as
\begin{spacing}{1.5}
	\qquad\qquad\quad$\beta = arg\mathop{min}\limits_{1\leq u\leq r}\{\widetilde{D}_{N}(P_{j},E_{u})\}$
\end{spacing}
\begin{spacing}{1.5}
	\qquad\qquad\quad\qquad\qquad\quad	$E_{u} \leftarrow P_{\beta}$
\end{spacing}

In order to make the process of classifying more straightforward, a flow chart is offered as follows.
Except for that, the corresponding pseudocode is given in \textbf{Algorithm 1}. 
\begin{algorithm}
	\caption{The details of the proposed algorithm } 
	\textbf{Input:} The sets of every pattern $P = \{P_{1},P_{2},P_{3},...,P_{k}\}$\\ 
	The sets of every sample $E = \{E_{1}, E_{2}, E_{3},...,E_{r}\}$\\
	\textbf{Output:} The results of classification of samples $E_{u}$\\
	$\textbf{for} \ j =1; \ j \leq k$ \ \textbf{do}  \\
	----- $\textbf{for} \ u =1; \ u \leq r$ \ \textbf{do}\\
	-----------Generate the distance $\widetilde{D}_{N}(P_{j},E_{u})$ between different PFSs by using the proposed method\\
	----- \textbf{end} \ \ \ \ \  /- Step 1 -/\\
	Choose the minimum value of $\widetilde{D}_{N}(P_{j},E_{u})$ as the final distance
	\ \ \ \ \  /- Step 2 -/\\
	Classify the tested sample $E_{u}$ into the corresponding pattern $P_{j}$	\ \ \ \ \  /- Step 3 -/\\
	\textbf{end}
\end{algorithm}

To clarify the specific process of the algorithm, an example is offered to illustrate details in handling data. Assume there are three medical patterns $P_{1}$, $P_{2}$ and $P_{3}$ which are expressed in the form of PFS in the finite universe of discourse $X = \{x_{1},x_{2},x_{3}\}$ and the details of the three medical patterns are written as follows:
\begin{spacing}{1.2}
	$\begin{aligned}
		&\qquad P_{1} = \{\langle x_{1},0.4, 0.7\rangle,\langle x_{2},0.5, 0.6\rangle,\langle x_{2},0.7, 0.4\rangle\}\\
		&\qquad P_{2} = \{\langle x_{1},0.6, 0.7\rangle,\langle x_{2},0.4, 0.5\rangle,\langle x_{2},0.9, 0.1\rangle\}\\
		&\qquad P_{3} = \{\langle x_{1},0.3, 0.6\rangle,\langle x_{2},0.7, 0.3\rangle,\langle x_{2},0.2, 0.8\rangle\}
	\end{aligned}$
\end{spacing}

Besides, two medical samples $S_{1}$ and $S_{2}$ are expressed in the form of PFS in the finite universe of discourse and defined as:
\begin{spacing}{1.2}
	$\begin{aligned}
		&\qquad S_{1} = \{\langle x_{1},0.6, 0.5\rangle,\langle x_{2},0.3, 0.8\rangle,\langle x_{2},0.2, 0.7\rangle\}\\
		&\qquad S_{2} = \{\langle x_{1},0.2, 0.6\rangle,\langle x_{2},0.7, 0.6\rangle,\langle x_{2},0.8, 0.3\rangle\}
	\end{aligned}$
\end{spacing}

What should be done is to categorize sample $S_{1}$ and $S_{2}$ into coincident classes. According to the procedure introduced above, the process of achieving the final outcomes is written as:

\textbf{1}: Generate the distance among  $P_{1}$, $P_{2}$, $P_{3}$ and $S_{1}$, $S_{2}$ by using the proposed method respectively. The results are written as:
\begin{spacing}{1.5}
	$\begin{aligned}
		&{D}_{N}(P_{1},S_{1}) = 0.4470, \widetilde{D}_{N}(P_{2},S_{1}) = 0.5134, \widetilde{D}_{N}(P_{3},S_{1}) = 0.4511\\
		&\widetilde{D}_{N}(P_{1},S_{2}) = 0.2076, \widetilde{D}_{N}(P_{2},S_{2}) = 0.3294, \widetilde{D}_{N}(P_{3},S_{2}) = 0.5127
	\end{aligned}$
\end{spacing}

\textbf{2}: Choose the smallest value of the distances generated by the proposed method. And according to the regulation, the process is written as:
\begin{spacing}{1.5}
	$\begin{aligned}
		&\qquad \qquad \quad \widetilde{D}_{N}^{chosen} = \widetilde{D}_{N}(P_{1},S_{1}) = 0.4470\\
		&\qquad \qquad \quad \widetilde{D}_{N}^{chosen}= \widetilde{D}_{N}(P_{1},S_{2}) = 0.2076
	\end{aligned}$
\end{spacing}

\textbf{3}: Classify the specific samples into corresponding patterns:
\begin{spacing}{1.5}
	\qquad\qquad\qquad\qquad$S_{1} \leftarrow P_{1}$,	$S_{2} \leftarrow P_{1}$
\end{spacing}

This is the full process of the designed algorithm.

\subsection{Application to COVID-19 Recognition}
In early 2020, the world was engulfed by a vast pandemic triggered by a novel strain of Coronavirus Disease-2019 (COVID-19), leading to numerous fatalities due to respiratory failure and various complications. This virus, known for its heightened contagiousness, has a more severe impact on health compared to other viral infections. As a result, promptly detecting those infected and ensuring their isolation became crucial in managing and mitigating the spread of the outbreak. Suppose there are four patients, namely $\mathcal{V}_1$, $\mathcal{V}_2$, $\mathcal{V}_3$ and $\mathcal{V}_4$, who are denoted as $\mathcal{V} = \{\mathcal{V}_1, \mathcal{V}_2, \mathcal{V}_3, \mathcal{V}_4\}$. Moreover, five attributes are symptoms given as $fever$, $cough$, $fatigue$, $trouble\ breathing$ and $sore\ throat$, which are denoted as $A = \{a_1, a_2, a_3, a_4, a_5\}$. The symptoms expressed in the form of PFS of patients in application on COVID-19 Recognition are provided in Table \ref{table9}.

\begin{table*}[t]
	\caption{Symptoms expressed in the form of PFS of patients in application on COVID-19 Recognition}
	\centering
	\begin{spacing}{1.20}
		\setlength{\tabcolsep}{3mm}{
			\begin{tabular}{cccccccccc}
				\midrule
				$Patients$ & $Symptom 1$ & $Symptom 2$ & $Symptom 3$ & $Symptom 4$& $Symptom 5$ \\
				\midrule
				$\mathcal{V}_{1}$ & $\langle a_{1},0.53, 0.70\rangle$ & $\langle a_{2},0.63, 0.45\rangle$ & $\langle a_{3},0.45, 0.56\rangle$ & $\langle a_{4},0.42, 0.67\rangle$ & $\langle a_{5},0.55, 0.61\rangle$ \\
				$\mathcal{V}_{2}$ & $\langle a_{1},0.92, 0.15\rangle$ & $\langle a_{2},0.86, 0.23\rangle$ & $\langle a_{3},0.75, 0.26\rangle$ & $\langle a_{4},0.77, 0.13\rangle$ & $\langle a_{5},0.93, 0.20\rangle$\\
				$\mathcal{V}_{3}$ & $\langle a_{1},0.67, 0.47\rangle$ & $\langle a_{2},0.89, 0.13\rangle$ & $\langle a_{3},0.68, 0.30\rangle$ & $\langle a_{4},0.81, 0.23\rangle$ & $\langle a_{5},0.88, 0.29\rangle$\\
				$\mathcal{V}_{4}$ & $\langle a_{1},0.66, 0.45\rangle$ & $\langle a_{2},0.75, 0.23\rangle$ & $\langle a_{3},0.56, 0.87\rangle$ & $\langle a_{4},0.57, 0.62\rangle$ & $\langle a_{5},0.61, 0.43\rangle$ \\
				\hline
		\end{tabular}}
	\end{spacing}
	\label{table9}
\end{table*}

\begin{table*}[h]
	\centering
	\caption{Judgements generated by different methods in application on COVID-19 Recognition}
	\begin{spacing}{1.20}
		\setlength{\tabcolsep}{6mm}{
			\begin{tabular}{l c c c c c}\hline
				$Methods$ & $Score (\mathcal{V}_{1}, \mathcal{V}_{2}, \mathcal{V}_{3}, \mathcal{V}_{4}$) &$Ranking$\\\hline
				PFWA \cite{DBLP:journals/ijis/Garg17}&0.7961, 0.9990,  0.9952, 0.9276  &  $\mathcal{V}_{2} > \mathcal{V}_{3} > \mathcal{V}_{4} > \mathcal{V}_{1}$\\
				PFWG \cite{DBLP:journals/ijis/Garg17}&-0.9022, -0.048, -0.3119, -0.8986 &  $\mathcal{V}_{2} > \mathcal{V}_{3} > \mathcal{V}_{4} > \mathcal{V}_{1}$\\
				$q$-ROFWA \cite{DBLP:journals/ijis/LiuW18} $(q=3)$ & 0.6679, 0.9967, 0.9859, 0.8512    & $\mathcal{V}_{2} > \mathcal{V}_{3} > \mathcal{V}_{4} > \mathcal{V}_{1}$ \\
				$q$-ROFWG \cite{DBLP:journals/ijis/LiuW18} $(q=3)$ & -0.8124, 0.0585, -0.2130, -0.8417 & $\mathcal{V}_{2} > \mathcal{V}_{3} > \mathcal{V}_{1} > \mathcal{V}_{4}$ \\
				$q$-ROFMM $(q=3)$ \cite{wang2019some} & -0.0689, 0.7183, 0.5929, 0.1880 & $\mathcal{V}_{2} > \mathcal{V}_{3} > \mathcal{V}_{4} > \mathcal{V}_{1}$ \\
				$q$-ROFDMM $q=3$ \cite{wang2019some} & -0.1175, 0.6679, 0.5025, -0.0200 &  $\mathcal{V}_{2} > \mathcal{V}_{3} > \mathcal{V}_{4} > \mathcal{V}_{1}$ \\
				Nan et al. v1 \cite{DBLP:journals/ijfs/NanZH22} & 0.0956, 0.7944, 0.7070, 0.4049 & $\mathcal{V}_{2} > \mathcal{V}_{3} > \mathcal{V}_{4} > \mathcal{V}_{1}$ \\
				Nan et al. v2 \cite{DBLP:journals/ijfs/NanZH22} & 0.1417, 0.8297, 0.7411, 0.4445 & $\mathcal{V}_{2} > \mathcal{V}_{3} > \mathcal{V}_{4} > \mathcal{V}_{1}$ \\\hline
				$Methods$ & $Distance (\mathcal{V}_{1}, \mathcal{V}_{2}, \mathcal{V}_{3}, \mathcal{V}_{4})$ &$Ranking$\\\hline
				Proposed method&0.6950, 0.1951, 0.2835, 0.5592 &$\mathcal{V}_{2} > \mathcal{V}_{3} > \mathcal{V}_{4} > \mathcal{V}_{1}$\\  \hline
		\end{tabular}}
	\end{spacing}
	\label{table10}
\end{table*}

Suppose a scenario where data pertaining to an individual exhibiting symptoms closely associated with the condition is denoted by PFS $\mathcal{V}_{ies}$ within A which is given as \cite{DBLP:journals/ijfs/NanZH22}:
\begin{spacing}{1.5}
	$\begin{aligned}
		&\mathcal{V}_{ies} = \{\langle x_{1},0.89, 0.18\rangle, \langle x_{2},0.91, 0.17\rangle, \langle x_{3},0.92, 0.10\rangle\},\\& \langle x_{4},0.97, 0.05\rangle, \langle x_{5},0.89, 0.19\rangle\}
	\end{aligned}
	$
\end{spacing}

\begin{table}[h]\small
	\centering
	\caption{Distances generated by proposed method in application on COVID-19 Recognition}
	\begin{spacing}{1.20}
		\begin{tabular}{c c c c c c}\hline
			$Patients$ &\multicolumn{5}{c}{$Distances$}\\\hline
			&$a_{1}$& $a_{2}$&$a_{3}$&$a_{4}$&$a_{5}$\\
			$\mathcal{V}_{1}$&$0.4951$&$0.2428$&$0.5742$&$0.6956$&$0.4075$\\
			$\mathcal{V}_{2}$&$0.0022$&$0.0068$&$0.0806$&$0.0966$&$0.0038$\\
			$\mathcal{V}_{3}$&$0.1738$&$0.0011$&$0.1615$&$0.0634$&$0.0021$\\
			$\mathcal{V}_{4}$&$0.1809$&$0.0708$&$0.6064$&$0.4605$&$0.2450$\\\hline
		\end{tabular}
	\end{spacing}
	\label{table11}
\end{table}

\begin{figure}
	\centering
	\includegraphics[width=0.5\linewidth]{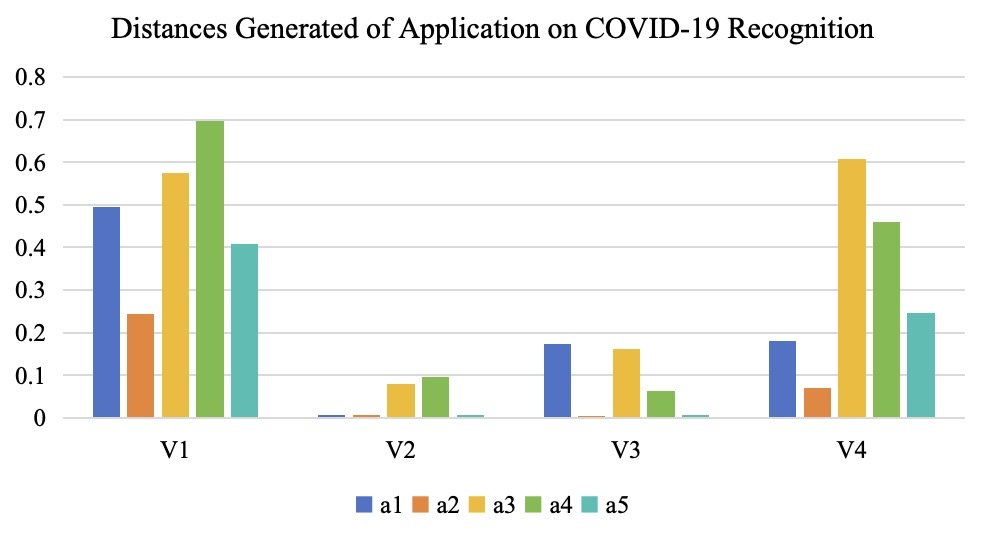}
	\caption{The visualized results of distances generated by proposed method in application on COVID-19 Recognition}
	\label{fig3}
\end{figure}

The objective of this task is to rank the likelihood of illness for four patients, that is, if a patient's condition is similar to the given examples of illness, then their probability of being ill is higher; if there is a significant difference from the examples of illness, then it is inclined to consider that the patient's probability of being ill is lower. Specifically, judgements generated by different methods and distances generated by proposed method in application on COVID-19  recognition are provided in Table \ref{table10} and \ref{table11}.  The method of comparison in the table involves scoring the probability of each patient being ill, with a higher score indicating a higher tendency towards illness. However, the method proposed in this article differs slightly; we measure the difference between the patient and the standard example. If the distance measurement obtained is small, it is considered that the patient has a higher probability of being ill; if the distance measurement is large, it is considered that the patient has a lower probability of being ill. Besides, the visualized results of distances generated by proposed method in application on COVID-19 Recognition is provided in Fig \ref{fig3}.

Overall, the solution proposed in this paper and the majority of the methods it was compared against achieved consistent results. Specifically, it was determined that Patient Two has the highest probability of being ill, followed by Patients Three, Four, and One, with a sequential decrease in the predicted probability of illness. This indicates the effectiveness of the distance measurement formula for Pythagorean fuzzy sets introduced in this study, as well as the validity of the newly designed algorithm.

\subsection{Applications to Symptom Diagnosis}
In this section, the medical pattern recognition problems are provided to further verify the validity of the proposed distance and corresponding algorithm.
\begin{table*}[t]
	\caption{Symptoms expressed in the form of PFS of patients in application 1}
	\centering
	\begin{spacing}{1.20}
		\setlength{\tabcolsep}{3mm}{
			\begin{tabular}{cccccccccc}
				\midrule
				$Patients$ & $Symptom 1$ & $Symptom 2$ & $Symptom 3$ & $Symptom 4$& $Symptom 5$ \\
				\midrule
				$P_{1}$ & $\langle a_{1},0.90, 0.10\rangle$ & $\langle a_{2},0.70, 0.20\rangle$ & $\langle a_{3},0.20, 0.80\rangle$ & $\langle a_{4},0.70, 0.20\rangle$ & $\langle a_{5},0.20, 0.70\rangle$ \\
				$P_{2}$ & $\langle a_{1},0.00, 0.70\rangle$ & $\langle a_{2},0.40, 0.50\rangle$ & $\langle a_{3},0.60, 0.20\rangle$ & $\langle a_{4},0.20, 0.70\rangle$ & $\langle a_{5},0.10, 0.20\rangle$\\
				$P_{3}$ & $\langle a_{1},0.70, 0.10\rangle$ & $\langle a_{2},0.70, 0.10\rangle$ & $\langle a_{3},0.00, 0.50\rangle$ & $\langle a_{4},0.10, 0.70\rangle$ & $\langle a_{5},0.00, 0.60\rangle$\\
				$P_{4}$ & $\langle a_{1},0.50, 0.10\rangle$ & $\langle a_{2},0.40, 0.30\rangle$ & $\langle a_{3},0.40, 0.50\rangle$ & $\langle a_{4},0.80, 0.20\rangle$ & $\langle a_{5},0.30, 0.40\rangle$ \\
				\hline
		\end{tabular}}
	\end{spacing}
	\label{table12}
\end{table*}
\begin{table*}[t]
	\caption{Symptoms expressed in the form of PFS of diagnoses in application 1}
	\centering
	\begin{spacing}{1.20}
		\setlength{\tabcolsep}{3mm}{
			\begin{tabular}{cccccccccc}
				\midrule
				$Diagnoses$ & $Symptom 1$ & $Symptom 2$ & $Symptom 3$ & $Symptom 4$& $Symptom 5$ \\
				\midrule
				$D_{1}$ & $\langle a_{1},0.30, 0.00\rangle$ & $\langle a_{2},0.30, 0.50\rangle$ & $\langle a_{3},0.20, 0.80\rangle$ & $\langle a_{4},0.70, 0.30\rangle$ & $\langle a_{5},0.20, 0.60\rangle$ \\
				$D_{2}$ & $\langle a_{1},0.00, 0.60\rangle$ & $\langle a_{2},0.20, 0.60\rangle$ & $\langle a_{3},0.00, 0.80\rangle$ & $\langle a_{4},0.50, 0.00\rangle$ & $\langle a_{5},0.10, 0.80\rangle$\\
				$D_{3}$ & $\langle a_{1},0.20, 0.20\rangle$ & $\langle a_{2},0.50, 0.20\rangle$ & $\langle a_{3},0.10, 0.70\rangle$ & $\langle a_{4},0.20, 0.60\rangle$ & $\langle a_{5},0.20, 0.80\rangle$\\
				$D_{4}$ & $\langle a_{1},0.20, 0.80\rangle$ & $\langle a_{2},0.10, 0.50\rangle$ & $\langle a_{3},0.70, 0.00\rangle$ & $\langle a_{4},0.10, 0.70\rangle$ & $\langle a_{5},0.20, 0.70\rangle$ \\
				$D_{5}$ & $\langle a_{1},0.20, 0.80\rangle$ & $\langle a_{2},0.00, 0.70\rangle$ & $\langle a_{3},0.20, 0.80\rangle$ & $\langle a_{4},0.10, 0.80\rangle$ & $\langle a_{5},0.80, 0.10\rangle$ \\
				\hline
		\end{tabular}}
	\end{spacing}
	\label{table13}
\end{table*}

\begin{table}[h]\small
	\centering
	\caption{Distances generated by proposed method in application 1}
	\begin{spacing}{1.20}
		\begin{tabular}{c c c c c c}\hline
			$Patients$ &\multicolumn{5}{c}{$Distances$}\\\hline
			&$D_{1}$& $D_{2}$&$D_{3}$&$D_{4}$&$D_{5}$\\
			$P_{1}$&$0.1930$&$0.3979$&$0.2707$&$0.6404$&$0.6942$\\
			$P_{2}$&$0.4230$&$0.3648$&$0.2882$&$0.1181$&$0.3191$\\
			$P_{3}$&$0.3558$&$0.4174$&$0.0996$&$0.4498$&$0.5463$\\
			$P_{4}$&$0.1658$&$0.2727$&$0.3270$&$0.4664$&$0.5125$\\\hline
		\end{tabular}
	\end{spacing}
	\label{table14}
\end{table}
\begin{table}[h]\scriptsize
	\centering
	\caption{Judgements generated by different methods in application 1}
	\begin{spacing}{1.20}
		\setlength{\tabcolsep}{0.6mm}{
			\begin{tabular}{l c c c c c}\hline
				$Methods$ &\multicolumn{4}{c}{$Diagnoses$}\\\hline
				&$P_{1}$& $P_{2}$&$P_{3}$&$P_{4}$\\
				Smuel and Rajakumar \cite{article5555}&$Stress$&$Spinal \ problem$&$Vision \ problem$&$Stress$\\
				Xiao \cite{article9999}&$Stress$&$Spinal \ problem$&$Vision \ problem$&$Stress$\\
				Proposed method&$Stress$&$Spinal \ problem$&$Vision \ problem$&$Stress$\\\hline
		\end{tabular}}
	\end{spacing}
	\label{table15}
\end{table}

\begin{table*}[t]\small
	\caption{Symptoms expressed in the form of PFS of patients in application 2}
	\centering
	\begin{spacing}{1.20}
		\setlength{\tabcolsep}{3mm}{
			\begin{tabular}{cccccccccc}
				\midrule
				$Patients$ & $Symptom 1$ & $Symptom 2$ & $Symptom 3$ & $Symptom 4$& $Symptom 5$ \\
				\midrule
				$P_{1}$ & $\langle a_{1},0.80, 0.10\rangle$ & $\langle a_{2},0.60, 0.10\rangle$ & $\langle a_{3},0.20, 0.80\rangle$ & $\langle a_{4},0.60, 0.10\rangle$ & $\langle a_{5},0.10, 0.60\rangle$ \\
				$P_{2}$ & $\langle a_{1},0.00, 0.80\rangle$ & $\langle a_{2},0.40, 0.40\rangle$ & $\langle a_{3},0.60, 0.10\rangle$ & $\langle a_{4},0.10, 0.70\rangle$ & $\langle a_{5},0.10, 0.80\rangle$\\
				$P_{3}$ & $\langle a_{1},0.80, 0.10\rangle$ & $\langle a_{2},0.80, 0.10\rangle$ & $\langle a_{3},0.00, 0.60\rangle$ & $\langle a_{4},0.20, 0.70\rangle$ & $\langle a_{5},0.00, 0.50\rangle$\\
				$P_{4}$ & $\langle a_{1},0.60, 0.10\rangle$ & $\langle a_{2},0.50, 0.40\rangle$ & $\langle a_{3},0.30, 0.40\rangle$ & $\langle a_{4},0.70, 0.20\rangle$ & $\langle a_{5},0.30, 0.40\rangle$ \\
				\hline
		\end{tabular}}
	\end{spacing}
		\label{table16}
\end{table*}
\begin{table*}[t]\small
	\caption{Symptoms expressed in the form of PFS of diagnoses in application 2}
	\centering
	\begin{spacing}{1.20}
		\setlength{\tabcolsep}{3mm}{
			\begin{tabular}{cccccccccc}
				\midrule
				$Diagnoses$ & $Symptom 1$ & $Symptom 2$ & $Symptom 3$ & $Symptom 4$& $Symptom 5$ \\
				\midrule
				$D_{1}$ & $\langle a_{1},0.40, 0.00\rangle$ & $\langle a_{2},0.30, 0.50\rangle$ & $\langle a_{3},0.10, 0.70\rangle$ & $\langle a_{4},0.40, 0.30\rangle$ & $\langle a_{5},0.10, 0.70\rangle$ \\
				$D_{2}$ & $\langle a_{1},0.70, 0.00\rangle$ & $\langle a_{2},0.20, 0.60\rangle$ & $\langle a_{3},0.00, 0.90\rangle$ & $\langle a_{4},0.70, 0.00\rangle$ & $\langle a_{5},0.10, 0.80\rangle$\\
				$D_{3}$ & $\langle a_{1},0.30, 0.30\rangle$ & $\langle a_{2},0.60, 0.10\rangle$ & $\langle a_{3},0.20, 0.70\rangle$ & $\langle a_{4},0.20, 0.60\rangle$ & $\langle a_{5},0.10, 0.90\rangle$\\
				$D_{4}$ & $\langle a_{1},0.10, 0.70\rangle$ & $\langle a_{2},0.20, 0.40\rangle$ & $\langle a_{3},0.80, 0.00\rangle$ & $\langle a_{4},0.20, 0.70\rangle$ & $\langle a_{5},0.20, 0.70\rangle$ \\
				$D_{5}$ & $\langle a_{1},0.10, 0.80\rangle$ & $\langle a_{2},0.00, 0.80\rangle$ & $\langle a_{3},0.20, 0.80\rangle$ & $\langle a_{4},0.20, 0.80\rangle$ & $\langle a_{5},0.80, 0.10\rangle$ \\
				\hline
		\end{tabular}}
	\end{spacing}
		\label{table17}
\end{table*}
\begin{table}[h]\small
	\centering
	\caption{Distances generated by proposed method in application 2}
	\begin{spacing}{1.20}
		\setlength{\tabcolsep}{1.4mm}{\begin{tabular}{c c c c c c}\hline
				$Patients$ &\multicolumn{5}{c}{$Distances$}\\\hline
				&$D_{1}$& $D_{2}$&$D_{3}$&$D_{4}$&$D_{5}$\\
				$P_{1}$&$0.2698$&$0.2235$&$0.2446$&$0.5878$&$0.6324$\\
				$P_{2}$&$0.2486$&$0.2143$&$0.2979$&$0.0673$&$0.5051$\\
				$P_{3}$&$0.2605$&$0.4448$&$0.1324$&$0.4853$&$0.5975$\\
				$P_{4}$&$0.1727$&$0.2075$&$0.3118$&$0.5085$&$0.5256$\\\hline
		\end{tabular}}
	\end{spacing}
		\label{table18}
\end{table}
\begin{table}[htbp]\scriptsize
	\centering
	\caption{Judgements generated by different methods in application 2}
	\begin{spacing}{1.20}
		\setlength{\tabcolsep}{1.1mm}{
			\begin{tabular}{l c c c c c}\hline
				$Methods$ &\multicolumn{4}{c}{$Diagnoses$}\\\hline
				&$P_{1}$& $P_{2}$&$P_{3}$&$P_{4}$\\\hline
				De et al. \cite{DE2001209}&$Malaria$&$Stomach\ problem$&$Typhoid$&$Malaria$\\
				Own \cite{article}&$Malaria$&$Stomach\ problem$&$Malaria$&$Malaria$\\
				Xiao \cite{article9999}&$Malaria$&$Stomach\ problem$&$Typhoid$&$Viral\ fever$\\
				Szmidt et al. \cite{10.1007/3-540-45718-6_30}&$Malaria$&$Stomach\ problem$&$Typhoid$&$Viral\ fever$\\
				Mondal et al. \cite{article1}&$Malaria$&$Stomach\ problem$&$Typhoid$&$Viral\ fever$\\
				Wei et al. \cite{WEI20114273}&$Malaria$&$Stomach\ problem$&$Typhoid$&$Viral\ fever$\\
				Pan et al. \cite{DBLP:journals/tfs/PanGDC22}&$Malaria$&$Stomach\ problem$&$Typhoid$&$Viral fever$\\
				Proposed method&$Malaria$&$Stomach\ problem$&$Typhoid$&$Viral\ fever$\\\hline
		\end{tabular}}
	\end{spacing}
		\label{table19}
\end{table}

\textbf{Application 1 :} Suppose there are four patients, namely $Ragu$, $Mathi$, $Velu$ and $Karthi$, who are denoted as $P = \{P_{1},P_{2},P_{3},P_{4} \}$. Besides, five attributes which are symptoms in fact are introduced as $Headache$, $Acidity$, $Burning\ eyes$,
$Back\ pain$ and $Depression$, which are denoted as $A = \{a_{1},a_{2},a_{3},a_{4},a_{5} \}$. More than that, diagnostic results are divided into five categories, $Stress$, $Ulcer$, $Vision\ problem$, $Spinal\ problem$ and $Blood\ pressure$, which are also denoted as $D = \{D_{1},D_{2},D_{3},D_{4}, D_{5}\}$. Specifically, symptoms expressed in the form of PFS of patient and diagnoses are provided in Table \ref{table12} and \ref{table13}. With the help of the concept of PFS, all of the examples are presented in the form of PFS, which is effective in judging proper patterns. And every details of the patients are shown in Table \ref{table14}. Additionally, the diagnoses in the form of PFS are also presented in Table \ref{table15}. 

\begin{figure}
	\centering
	\includegraphics[width=0.5\linewidth]{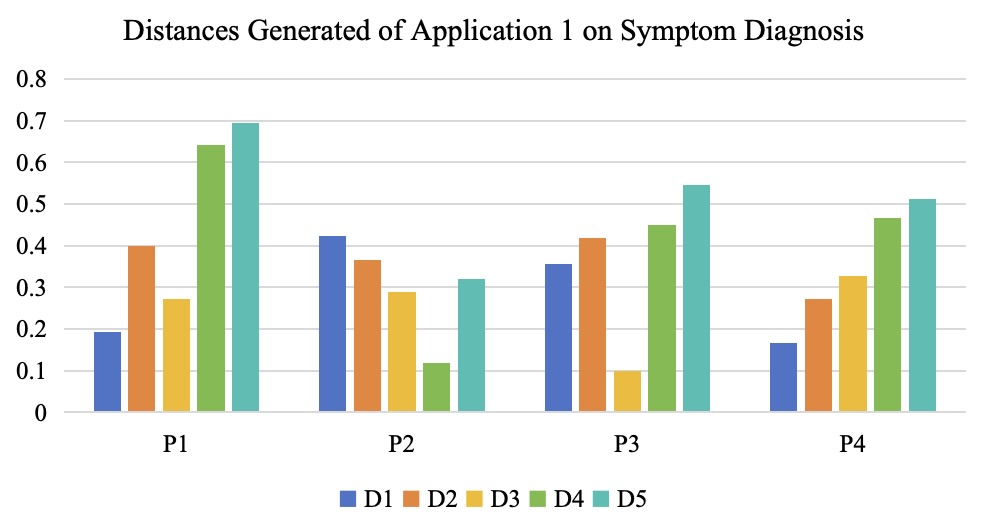}
	\caption{The visualized results of distances generated by proposed method in application 1 on Symptom Diagnosis}
	\label{fig4}
\end{figure}

By using the proposed method of measuring distance between PFSs, all the results generated by the proposed method are shown in table $16$.  According the standards raised in the algorithm and other methods, all of final judgements are presented in table $17$. Besides, the visualized results of distances generated by proposed method in application 1 on Symptom Diagnosis is provided in Fig \ref{fig4}. After checking the results generated by the proposed methods, $P_{1}$ has the least value of $\widetilde{D}_{N} = 0.1930$; $P_{2}$ has the least value of $\widetilde{D}_{N} = 0.1181$; $P_{3}$ has the least value of $\widetilde{D}_{N} = 0.0996$; $P_{4}$ has the least value of $\widetilde{D}_{N} = 0.1658$. All in all, the results produced by Sumuel and Rajakumar's method and Xiao's method conform to the ones produced by the proposed method, which proves that the accuracy and validity of the proposed method in real application and the feasibility of the proposed method in practical usage.

\begin{figure}
	\centering
	\includegraphics[width=0.5\linewidth]{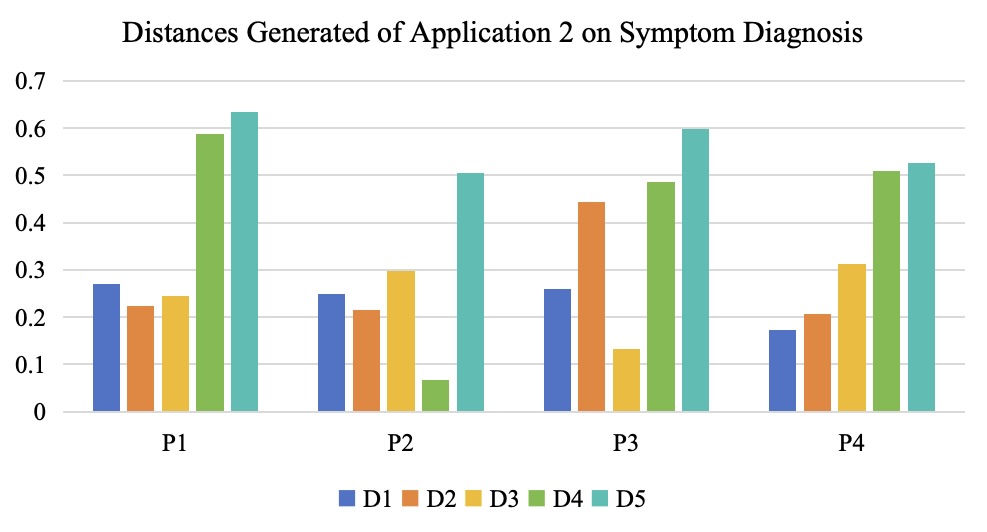}
	\caption{The visualized results of distances generated by proposed method in application 2 on Symptom Diagnosis}
	\label{fig5}
\end{figure}

\textbf{Application 2 :} Suppose there are four patients, namely $Al$, $Bob$, $Joe$ and $Ted$, who are denoted as $P = \{P_{1},P_{2},P_{3},P_{4} \}$. Besides, five attributes which are symptoms in fact are introduced as $Temperature$, $Headache$, $Stomach\ pain$,
$Cough$ and $Chest\ pain$, which are denoted as $A = \{a_{1},a_{2},a_{3},a_{4},a_{5} \}$. More than that, diagnostic results are divided into five categories, $Viral\ fever$, $Malaria$, $Typhoid$, $Stomach\ problem$ and $Chest$, which are also denoted as $D = \{D_{1},D_{2},D_{3},D_{4}, D_{5}\}$. Specifically, symptoms expressed in the form of PFS of patient and diagnoses are provided in Table \ref{table16} and \ref{table17}. With the help of the concept of PFS, all of the examples are presented in the form of PFS, which is effective in judging proper patterns. And every details of the patients are shown in Table \ref{table18}. Additionally, the diagnoses in the form of PFS are also presented in Table \ref{table19} and the visualized results of distances generated by proposed method in application 2 on Symptom Diagnosis is provided in Fig \ref{fig5}. 

By using the proposed method of measuring distance between PFSs, all the results generated by the proposed method are shown in table $8$. According the standards raised in the algorithm, all of final judgements are presented in table $9$. It can be concluded that $Al\ (P_{1})$ is diagnosed that he suffers from $Malaria\ (D_{2})$, $Bob\ (P_{2})$ is diagnosed that he suffers from $Stomach \ problem\ (D_{4})$, $Joe\ (P_{3})$ is diagnosed that she suffers from $Typhoid\ (D_{3})$ and $Ted\ (P_{4})$ is diagnosed that he suffers from $Viral \ fever (D_{1})$. And all the judgements produced by other methods and the results generated by the proposed method are placed together in table $9$ to verify the correctness of the latter one. In the chart, what is the most obvious is that all of the methods have reach an agreement that $Al\ (P_{1})$ is diagnosed with $Malaria\ (D_{2})$ and $Bob\ (P_{2})$ is diagnosed with $Stomach \ problem\ (D_{4})$, which is satisfying and rational. However, when coming to judge the situation of patient $Joe\ (P_{3})$, the diagnosis vary among different methods. Five of them give judgements that $Joe\ (P_{3})$ is suffering from $Typhoid \ (D_{3})$ and only one of them considers that this patient is suffering from $Malaria\ (D_{2})$. Additionally, with respect to $Ted\ (P_{4})$, four of the methods choose $Viral \ fever (D_{1})$ as the diagnosis of $Ted\ (P_{4})$ and the other two methods regard that $Ted\ (P_{4})$ is suffering from $Malaria\ (D_{2})$. All of the results demonstrate that it is very difficult to diagnose $Ted\ (P_{4})$, because there may be a potential relationship between $Viral \ fever (D_{1})$ and $Malaria\ (D_{2})$ leading to a conflicting stage when comparing the results of different methods. Anyway, the results produced by Szmidt et al.'s method, Mondal et al.'s method and Wei et al.'s method conform to the ones produced by the proposed method, which proves that the accuracy and validity of the proposed method in real application and the feasibility of the proposed method in practical usage.

\section{Conclusion}
In the theory of pythagorean fuzzy sets, how to accurately and properly measure the distance between PFSs is still an open issue, which may lead to chaos in pattern recognition. To solve this problem, in this paper, a completely new method of distance measure between different PFSs is proposed, which satisfy all of the properties required by the axiom of distance measurement. The main advantage of the proposed
method is that it considers the index of hesitance and distribute its mass to membership and non-membership in a reasonable way to strengthen the role which membership and non-membership plays in generating distances between PFSs. Besides, the reduction in the index of hesitance is also very important in alleviating the vagueness in PFSs and helpful in recognizing corresponding targets or patterns. All in all, the proposed method produces much more rational results than previous methods, which is more closer to actual situations and conforms to intuitive judgements. Moreover, the algorithm which is developed on the basis of the proposed method offers a promising and reliable solution to address the recognition problems in pattern recognition applications. In future work, we will consider designing a PFS distance based on a learnable matrix, which can better utilize trainable data to set the values of matrix elements. This will enhance the distance measurement performance, allowing for improved accuracy in a broader range of applications.


\bibliographystyle{elsarticle-num}
\bibliography{cite2}

\begin{thebibliography}{10}
\expandafter\ifx\csname url\endcsname\relax
  \def\url#1{\texttt{#1}}\fi
\expandafter\ifx\csname urlprefix\endcsname\relax\def\urlprefix{URL }\fi
\expandafter\ifx\csname href\endcsname\relax
  \def\href#1#2{#2} \def\path#1{#1}\fi

\bibitem{Deng2020ScienceChina}
Y.~Deng, Uncertainty measure in evidence theory, SCIENCE CHINA Information
  Sciences 64 (2021) 10.1007/s11432--020--3006--9.

\bibitem{DBLP:journals/tfs/WuZYXLW24}
C.~Wu, Q.~Zhang, L.~Yin, Q.~Xie, N.~Luo, G.~Wang,
  \href{https://doi.org/10.1109/TFUZZ.2023.3287834}{Data-driven interval
  granulation approach based on uncertainty principle for efficient
  classification}, {IEEE} Trans. Fuzzy Syst. 32~(1) (2024) 12--26.
\newblock \href {https://doi.org/10.1109/TFUZZ.2023.3287834}
  {\path{doi:10.1109/TFUZZ.2023.3287834}}.
\newline\urlprefix\url{https://doi.org/10.1109/TFUZZ.2023.3287834}

\bibitem{DBLP:journals/tit/ChenL24}
G.~Chen, S.~C. Liew, \href{https://doi.org/10.1109/TIT.2023.3315459}{An index
  policy for minimizing the uncertainty-of-information of markov sources},
  {IEEE} Trans. Inf. Theory 70~(1) (2024) 698--721.
\newblock \href {https://doi.org/10.1109/TIT.2023.3315459}
  {\path{doi:10.1109/TIT.2023.3315459}}.
\newline\urlprefix\url{https://doi.org/10.1109/TIT.2023.3315459}

\bibitem{DBLP:journals/isci/HeD23}
Y.~He, Y.~Deng, \href{https://doi.org/10.1016/j.ins.2022.11.036}{{TDQMF:}
  two-dimensional quantum mass function}, Inf. Sci. 621 (2023) 749--765.
\newblock \href {https://doi.org/10.1016/J.INS.2022.11.036}
  {\path{doi:10.1016/J.INS.2022.11.036}}.
\newline\urlprefix\url{https://doi.org/10.1016/j.ins.2022.11.036}

\bibitem{DBLP:journals/inffus/LiuLXS23}
X.~Liu, S.~Liu, J.~Xiang, R.~Sun,
  \href{https://doi.org/10.1016/j.inffus.2023.01.009}{A conflict evidence
  fusion method based on the composite discount factor and the game theory},
  Inf. Fusion 94 (2023) 1--16.
\newblock \href {https://doi.org/10.1016/J.INFFUS.2023.01.009}
  {\path{doi:10.1016/J.INFFUS.2023.01.009}}.
\newline\urlprefix\url{https://doi.org/10.1016/j.inffus.2023.01.009}

\bibitem{DBLP:journals/tcyb/HuangLD23}
L.~Huang, Z.~Liu, J.~Dezert,
  \href{https://doi.org/10.1109/TCYB.2021.3133890}{Cross-domain pattern
  classification with distribution adaptation based on evidence theory}, {IEEE}
  Trans. Cybern. 53~(2) (2023) 718--731.
\newblock \href {https://doi.org/10.1109/TCYB.2021.3133890}
  {\path{doi:10.1109/TCYB.2021.3133890}}.
\newline\urlprefix\url{https://doi.org/10.1109/TCYB.2021.3133890}

\bibitem{DBLP:journals/tsmc/CuiD23}
Y.~Cui, X.~Deng, \href{https://doi.org/10.1109/TSMC.2022.3233156}{Plausibility
  entropy: {A} new total uncertainty measure in evidence theory based on
  plausibility function}, {IEEE} Trans. Syst. Man Cybern. Syst. 53~(6) (2023)
  3833--3844.
\newblock \href {https://doi.org/10.1109/TSMC.2022.3233156}
  {\path{doi:10.1109/TSMC.2022.3233156}}.
\newline\urlprefix\url{https://doi.org/10.1109/TSMC.2022.3233156}

\bibitem{DBLP:journals/tcyb/ZhouDY24}
Q.~Zhou, Y.~Deng, R.~R. Yager,
  \href{https://doi.org/10.1109/TCYB.2023.3295179}{{CD-BFT:} canonical
  decomposition-based belief functions transformation in possibility theory},
  {IEEE} Trans. Cybern. 54~(1) (2024) 611--623.
\newblock \href {https://doi.org/10.1109/TCYB.2023.3295179}
  {\path{doi:10.1109/TCYB.2023.3295179}}.
\newline\urlprefix\url{https://doi.org/10.1109/TCYB.2023.3295179}

\bibitem{DBLP:journals/apin/HeX22}
Y.~He, F.~Xiao, \href{https://doi.org/10.1007/s10489-021-02525-w}{A new base
  function in basic probability assignment for conflict management}, Appl.
  Intell. 52~(4) (2022) 4473--4487.
\newblock \href {https://doi.org/10.1007/S10489-021-02525-W}
  {\path{doi:10.1007/S10489-021-02525-W}}.
\newline\urlprefix\url{https://doi.org/10.1007/s10489-021-02525-w}

\bibitem{DBLP:journals/inffus/KangZ24}
B.~Kang, C.~Zhao, \href{https://doi.org/10.1016/j.inffus.2023.102102}{Deceptive
  evidence detection in information fusion of belief functions based on
  reinforcement learning}, Inf. Fusion 103 (2024) 102102.
\newblock \href {https://doi.org/10.1016/J.INFFUS.2023.102102}
  {\path{doi:10.1016/J.INFFUS.2023.102102}}.
\newline\urlprefix\url{https://doi.org/10.1016/j.inffus.2023.102102}

\bibitem{DBLP:journals/cam/HeD22}
Y.~He, Y.~Deng, \href{https://doi.org/10.1007/s40314-021-01697-y}{{MMGET:} a
  markov model for generalized evidence theory}, Comput. Appl. Math. 41~(1)
  (2022).
\newblock \href {https://doi.org/10.1007/S40314-021-01697-Y}
  {\path{doi:10.1007/S40314-021-01697-Y}}.
\newline\urlprefix\url{https://doi.org/10.1007/s40314-021-01697-y}

\bibitem{DBLP:journals/tsmc/HuangGSZJY23}
Y.~Huang, B.~Ge, J.~Sun, B.~Zhao, J.~Jiang, K.~Yang,
  \href{https://doi.org/10.1109/TSMC.2022.3186763}{Belief-based preference
  structure and elicitation in the graph model for conflict resolution}, {IEEE}
  Trans. Syst. Man Cybern. Syst. 53~(2) (2023) 727--740.
\newblock \href {https://doi.org/10.1109/TSMC.2022.3186763}
  {\path{doi:10.1109/TSMC.2022.3186763}}.
\newline\urlprefix\url{https://doi.org/10.1109/TSMC.2022.3186763}

\bibitem{Yager2019}
R.~R. Yager, Entailment for measure based belief structures, Information Fusion
  47 (2019) 111--116.

\bibitem{DBLP:journals/tfs/BhowalSYGS23}
P.~Bhowal, S.~Sen, J.~H. Yoon, Z.~W. Geem, R.~Sarkar,
  \href{https://doi.org/10.1109/TFUZZ.2022.3206504}{Evaluation of fuzzy
  measures using dempster-shafer belief structure: {A} classifier fusion
  framework}, {IEEE} Trans. Fuzzy Syst. 31~(5) (2023) 1593--1603.
\newblock \href {https://doi.org/10.1109/TFUZZ.2022.3206504}
  {\path{doi:10.1109/TFUZZ.2022.3206504}}.
\newline\urlprefix\url{https://doi.org/10.1109/TFUZZ.2022.3206504}

\bibitem{DBLP:journals/tit/MasihuddinM24}
Masihuddin, N.~Misra, \href{https://doi.org/10.1109/TIT.2023.3329701}{Shrinkage
  estimators dominating some naive estimators of the selected entropy}, {IEEE}
  Trans. Inf. Theory 70~(1) (2024) 532--549.
\newblock \href {https://doi.org/10.1109/TIT.2023.3329701}
  {\path{doi:10.1109/TIT.2023.3329701}}.
\newline\urlprefix\url{https://doi.org/10.1109/TIT.2023.3329701}

\bibitem{DBLP:journals/tie/Novak24}
Z.~Novak, \href{https://doi.org/10.1109/TIE.2023.3283690}{Confidence weighted
  learning entropy for fault-tolerant control of a {PMSM} with a
  high-resolution hall encoder}, {IEEE} Trans. Ind. Electron. 71~(5) (2024)
  5176--5186.
\newblock \href {https://doi.org/10.1109/TIE.2023.3283690}
  {\path{doi:10.1109/TIE.2023.3283690}}.
\newline\urlprefix\url{https://doi.org/10.1109/TIE.2023.3283690}

\bibitem{DBLP:journals/tit/WuLXH24}
C.~Wu, Y.~Li, E.~L. Xu, G.~Han,
  \href{https://doi.org/10.1109/TIT.2023.3318265}{R{\'{e}}nyi entropy rate of
  stationary ergodic processes}, {IEEE} Trans. Inf. Theory 70~(1) (2024) 1--15.
\newblock \href {https://doi.org/10.1109/TIT.2023.3318265}
  {\path{doi:10.1109/TIT.2023.3318265}}.
\newline\urlprefix\url{https://doi.org/10.1109/TIT.2023.3318265}

\bibitem{DBLP:journals/soco/HeD23}
Y.~He, Y.~Deng, \href{https://doi.org/10.1007/s00500-023-07947-x}{Ordinal
  belief entropy}, Soft Comput. 27~(11) (2023) 6973--6981.
\newblock \href {https://doi.org/10.1007/S00500-023-07947-X}
  {\path{doi:10.1007/S00500-023-07947-X}}.
\newline\urlprefix\url{https://doi.org/10.1007/s00500-023-07947-x}

\bibitem{Kang2019}
B.~Kang, P.~Zhang, Z.~Gao, G.~Chhipi-Shrestha, K.~Hewage, R.~Sadiq,
  Environmental assessment under uncertainty using {Dempster--Shafer} theory
  and {Z}-numbers, Journal of Ambient Intelligence and Humanized Computing
  (2019) DOI: 10.1007/s12652--019--01228--y.

\bibitem{dehshiri2024evaluation}
S.~J.~H. Dehshiri, M.~Amiri, Evaluation of blockchain implementation solutions
  in the sustainable supply chain: A novel hybrid decision approach based on
  z-numbers, Expert Systems with Applications 235 (2024) 121123.

\bibitem{li2020newuncertainty}
Y.~Li, H.~Garg, Y.~Deng, {A New Uncertainty Measure of Discrete Z-numbers},
  {International Journal of Fuzzy Systems} {22}~({3}) ({2020}) 760--776.

\bibitem{mandal2024failure}
P.~Mandal, S.~Samanta, M.~Pal, Failure mode and effects analysis in
  consensus-based gdm for surface-guided deep inspiration breath-hold breast
  radiotherapy for breast cancer under the framework of linguistic z-number,
  Information Sciences 658 (2024) 120016.

\bibitem{Liu2019b}
Q.~Liu, Y.~Tian, B.~Kang, {Derive knowledge of Z-number from the perspective of
  Dempster--Shafer evidence theory}, Engineering Applications of Artificial
  Intelligence 85 (2019) 754--764.

\bibitem{sotoudeh2024state}
A.~Sotoudeh-Anvari, A state-of-the-art review on d number (2012-2022): A
  scientometric analysis, Engineering Applications of Artificial Intelligence
  127 (2024) 107309.

\bibitem{xiao2019multiple}
F.~Xiao, {A multiple-criteria decision-making method based on D numbers and
  belief entropy}, International Journal of Fuzzy Systems 21~(4) (2019)
  1144--1153.

\bibitem{Deng2019}
X.~Deng, W.~Jiang, A total uncertainty measure for {D} numbers based on belief
  intervals, International Journal of Intelligent Systems 34~(12) (2019)
  3302--3316.

\bibitem{Zadeh1965}
L.~A. Zadeh, Fuzzy sets, Information and control 8~(3) (1965) 338--353.

\bibitem{Zadeh1979}
L.~A. Zadeh, Fuzzy sets and information granularity, Advances in fuzzy set
  theory and applications 11 (1979) 3--18.

\bibitem{DBLP:journals/isci/RaniCM24}
P.~Rani, S.~Chen, A.~R. Mishra,
  \href{https://doi.org/10.1016/j.ins.2023.119990}{Multi-attribute
  decision-making based on similarity measure between picture fuzzy sets and
  the {MARCOS} method}, Inf. Sci. 658 (2024) 119990.
\newblock \href {https://doi.org/10.1016/J.INS.2023.119990}
  {\path{doi:10.1016/J.INS.2023.119990}}.
\newline\urlprefix\url{https://doi.org/10.1016/j.ins.2023.119990}

\bibitem{DBLP:journals/tfs/0001DX023}
J.~Zhan, J.~Deng, Z.~Xu, L.~Mart{\'{\i}}nez,
  \href{https://doi.org/10.1109/TFUZZ.2023.3237646}{A three-way decision
  methodology with regret theory via triangular fuzzy numbers in incomplete
  multiscale decision information systems}, {IEEE} Trans. Fuzzy Syst. 31~(8)
  (2023) 2773--2787.
\newblock \href {https://doi.org/10.1109/TFUZZ.2023.3237646}
  {\path{doi:10.1109/TFUZZ.2023.3237646}}.
\newline\urlprefix\url{https://doi.org/10.1109/TFUZZ.2023.3237646}

\bibitem{he2022ordinal}
Y.~He, Y.~Deng, Ordinal fuzzy entropy, Iranian Journal of Fuzzy Systems 19~(3)
  (2022) 171--186.

\bibitem{Atanassov1999}
K.~T. Atanassov, Intuitionistic fuzzy sets, Springer, 1999, pp. 1--137.

\bibitem{DBLP:journals/isci/JebadassB24}
J.~R. Jebadass, P.~Balasubramaniam,
  \href{https://doi.org/10.1016/j.ins.2023.119811}{Color image enhancement
  technique based on interval-valued intuitionistic fuzzy set}, Inf. Sci. 653
  (2024) 119811.
\newblock \href {https://doi.org/10.1016/J.INS.2023.119811}
  {\path{doi:10.1016/J.INS.2023.119811}}.
\newline\urlprefix\url{https://doi.org/10.1016/j.ins.2023.119811}

\bibitem{DBLP:journals/eswa/DongW24}
J.~Dong, S.~Wan,
  \href{https://doi.org/10.1016/j.eswa.2023.121213}{Interval-valued
  intuitionistic fuzzy best-worst method with additive consistency}, Expert
  Syst. Appl. 236 (2024) 121213.
\newblock \href {https://doi.org/10.1016/J.ESWA.2023.121213}
  {\path{doi:10.1016/J.ESWA.2023.121213}}.
\newline\urlprefix\url{https://doi.org/10.1016/j.eswa.2023.121213}

\bibitem{DBLP:journals/inffus/TiwariZQM24}
P.~Tiwari, L.~Zhang, Z.~Qu, G.~Muhammad,
  \href{https://doi.org/10.1016/j.inffus.2023.102085}{Quantum fuzzy neural
  network for multimodal sentiment and sarcasm detection}, Inf. Fusion 103
  (2024) 102085.
\newblock \href {https://doi.org/10.1016/J.INFFUS.2023.102085}
  {\path{doi:10.1016/J.INFFUS.2023.102085}}.
\newline\urlprefix\url{https://doi.org/10.1016/j.inffus.2023.102085}

\bibitem{DBLP:journals/eswa/MartinoS22}
F.~D. Martino, S.~Sessa, \href{https://doi.org/10.1016/j.eswa.2021.116340}{A
  novel quantum inspired genetic algorithm to initialize cluster centers in
  fuzzy c-means}, Expert Syst. Appl. 191 (2022) 116340.
\newblock \href {https://doi.org/10.1016/J.ESWA.2021.116340}
  {\path{doi:10.1016/J.ESWA.2021.116340}}.
\newline\urlprefix\url{https://doi.org/10.1016/j.eswa.2021.116340}

\bibitem{Yager2014a}
R.~Yager, {Pythagorean Membership Grades in Multicriteria Decision Making},
  IEEE Transactions on Fuzzy Systems 22 (2014) 958--965.

\bibitem{DBLP:journals/ijis/LvZZLK22}
W.~Lv, S.~Zeng, J.~Zhou, T.~Li, A.~S.~V. Koe,
  \href{https://doi.org/10.1002/int.22849}{Interval-valued pythagorean fuzzy
  linguistic {KPCA} model based on {TOPSIS} and its application for emergency
  group decision making}, Int. J. Intell. Syst. 37~(9) (2022) 6415--6437.
\newblock \href {https://doi.org/10.1002/INT.22849}
  {\path{doi:10.1002/INT.22849}}.
\newline\urlprefix\url{https://doi.org/10.1002/int.22849}

\bibitem{DBLP:journals/eswa/AkramLA22}
M.~Akram, A.~Luqman, J.~C.~R. Alcantud,
  \href{https://doi.org/10.1016/j.eswa.2022.116945}{An integrated {ELECTRE-I}
  approach for risk evaluation with hesitant pythagorean fuzzy information},
  Expert Syst. Appl. 200 (2022) 116945.
\newblock \href {https://doi.org/10.1016/J.ESWA.2022.116945}
  {\path{doi:10.1016/J.ESWA.2022.116945}}.
\newline\urlprefix\url{https://doi.org/10.1016/j.eswa.2022.116945}

\bibitem{DBLP:journals/tfs/EjegwaWFZT22}
P.~A. Ejegwa, S.~Wen, Y.~Feng, W.~Zhang, N.~Tang,
  \href{https://doi.org/10.1109/TFUZZ.2021.3063794}{Novel pythagorean fuzzy
  correlation measures via pythagorean fuzzy deviation, variance, and
  covariance with applications to pattern recognition and career placement},
  {IEEE} Trans. Fuzzy Syst. 30~(6) (2022) 1660--1668.
\newblock \href {https://doi.org/10.1109/TFUZZ.2021.3063794}
  {\path{doi:10.1109/TFUZZ.2021.3063794}}.
\newline\urlprefix\url{https://doi.org/10.1109/TFUZZ.2021.3063794}

\bibitem{DBLP:journals/asc/SarkarCB23}
B.~Sarkar, D.~Chakraborty, A.~Biswas,
  \href{https://doi.org/10.1016/j.asoc.2023.110332}{Development of type-2
  pythagorean fuzzy set with its application to sustainable transport system
  selection}, Appl. Soft Comput. 142 (2023) 110332.
\newblock \href {https://doi.org/10.1016/J.ASOC.2023.110332}
  {\path{doi:10.1016/J.ASOC.2023.110332}}.
\newline\urlprefix\url{https://doi.org/10.1016/j.asoc.2023.110332}

\bibitem{SZMIDT2000505}
E.~Szmidt, J.~Kacprzyk,
  \href{http://www.sciencedirect.com/science/article/pii/S0165011498002449}{Distances
  between intuitionistic fuzzy sets}, Fuzzy Sets and Systems 114~(3) (2000) 505
  -- 518.
\newblock \href {https://doi.org/https://doi.org/10.1016/S0165-0114(98)00244-9}
  {\path{doi:https://doi.org/10.1016/S0165-0114(98)00244-9}}.
\newline\urlprefix\url{http://www.sciencedirect.com/science/article/pii/S0165011498002449}

\bibitem{GRZEGORZEWSKI2004319}
P.~Grzegorzewski,
  \href{http://www.sciencedirect.com/science/article/pii/S0165011403003543}{Distances
  between intuitionistic fuzzy sets and/or interval-valued fuzzy sets based on
  the hausdorff metric}, Fuzzy Sets and Systems 148~(2) (2004) 319 -- 328.
\newblock \href {https://doi.org/https://doi.org/10.1016/j.fss.2003.08.005}
  {\path{doi:https://doi.org/10.1016/j.fss.2003.08.005}}.
\newline\urlprefix\url{http://www.sciencedirect.com/science/article/pii/S0165011403003543}

\bibitem{article}
c.-m. Own, Switching between type-2 fuzzy sets and intuitionistic fuzzy sets:
  An application in medical diagnosis, Appl. Intell. 31 (2009) 283--291.
\newblock \href {https://doi.org/10.1007/s10489-008-0126-y}
  {\path{doi:10.1007/s10489-008-0126-y}}.

\bibitem{DE2001209}
S.~K. De, R.~Biswas, A.~R. Roy,
  \href{http://www.sciencedirect.com/science/article/pii/S0165011498002358}{An
  application of intuitionistic fuzzy sets in medical diagnosis}, Fuzzy Sets
  and Systems 117~(2) (2001) 209 -- 213.
\newblock \href {https://doi.org/https://doi.org/10.1016/S0165-0114(98)00235-8}
  {\path{doi:https://doi.org/10.1016/S0165-0114(98)00235-8}}.
\newline\urlprefix\url{http://www.sciencedirect.com/science/article/pii/S0165011498002358}

\bibitem{10.1007/3-540-45718-6_30}
E.~Szmidt, J.~Kacprzyk, Intuitionistic fuzzy sets in intelligent data analysis
  for medical diagnosis, in: V.~N. Alexandrov, J.~J. Dongarra, B.~A. Juliano,
  R.~S. Renner, C.~J.~K. Tan (Eds.), Computational Science - ICCS 2001,
  Springer Berlin Heidelberg, Berlin, Heidelberg, 2001, pp. 263--271.

\bibitem{WEI20114273}
C.-P. Wei, P.~Wang, Y.-Z. Zhang,
  \href{http://www.sciencedirect.com/science/article/pii/S0020025511002751}{Entropy,
  similarity measure of interval-valued intuitionistic fuzzy sets and their
  applications}, Information Sciences 181~(19) (2011) 4273 -- 4286.
\newblock \href {https://doi.org/https://doi.org/10.1016/j.ins.2011.06.001}
  {\path{doi:https://doi.org/10.1016/j.ins.2011.06.001}}.
\newline\urlprefix\url{http://www.sciencedirect.com/science/article/pii/S0020025511002751}

\bibitem{article1}
S.~Pramanik, K.~Mondal, Intuitionistic fuzzy similarity measure based on
  tangent function and its application to multi-attribute decision making,
  Global Journal of Advanced Research 2 (2015) 464--471.

\bibitem{CHEN2018129}
T.-Y. Chen,
  \href{http://www.sciencedirect.com/science/article/pii/S1566253517300763}{Remoteness
  index-based pythagorean fuzzy vikor methods with a generalized distance
  measure for multiple criteria decision analysis}, Information Fusion 41
  (2018) 129 -- 150.
\newblock \href {https://doi.org/https://doi.org/10.1016/j.inffus.2017.09.003}
  {\path{doi:https://doi.org/10.1016/j.inffus.2017.09.003}}.
\newline\urlprefix\url{http://www.sciencedirect.com/science/article/pii/S1566253517300763}

\bibitem{article5}
S.-M. Chen, J.-M. Tan, Handling multicriteria fuzzy decision-making problems
  based on vague set theory, Fuzzy Sets and Systems 67 (1994) 163--172.
\newblock \href {https://doi.org/10.1016/0165-0114(94)90084-1}
  {\path{doi:10.1016/0165-0114(94)90084-1}}.

\bibitem{DBLP:journals/ijis/Garg17}
H.~Garg, \href{https://doi.org/10.1002/int.21860}{Generalized pythagorean fuzzy
  geometric aggregation operators using einstein \emph{t}-norm and
  \emph{t}-conorm for multicriteria decision-making process}, Int. J. Intell.
  Syst. 32~(6) (2017) 597--630.
\newblock \href {https://doi.org/10.1002/INT.21860}
  {\path{doi:10.1002/INT.21860}}.
\newline\urlprefix\url{https://doi.org/10.1002/int.21860}

\bibitem{DBLP:journals/ijis/LiuW18}
P.~Liu, P.~Wang, \href{https://doi.org/10.1002/int.21927}{Some q-rung orthopair
  fuzzy aggregation operators and their applications to multiple-attribute
  decision making}, Int. J. Intell. Syst. 33~(2) (2018) 259--280.
\newblock \href {https://doi.org/10.1002/INT.21927}
  {\path{doi:10.1002/INT.21927}}.
\newline\urlprefix\url{https://doi.org/10.1002/int.21927}

\bibitem{wang2019some}
J.~Wang, R.~Zhang, X.~Zhu, Z.~Zhou, X.~Shang, W.~Li, Some q-rung orthopair
  fuzzy muirhead means with their application to multi-attribute group decision
  making, Journal of Intelligent \& Fuzzy Systems 36~(2) (2019) 1599--1614.

\bibitem{DBLP:journals/ijfs/NanZH22}
T.~Nan, H.~Zhang, Y.~He,
  \href{https://doi.org/10.1007/s40815-022-01261-8}{Pythagorean fuzzy full
  implication triple {I} method and its application in medical diagnosis}, Int.
  J. Fuzzy Syst. 24~(5) (2022) 2250--2263.
\newblock \href {https://doi.org/10.1007/S40815-022-01261-8}
  {\path{doi:10.1007/S40815-022-01261-8}}.
\newline\urlprefix\url{https://doi.org/10.1007/s40815-022-01261-8}

\bibitem{article5555}
X.~Deng, Analyzing the monotonicity of belief interval based uncertainty
  measures in belief function theory, International Journal of Intelligent
  Systems 33 (03 2018).
\newblock \href {https://doi.org/10.1002/int.21999}
  {\path{doi:10.1002/int.21999}}.

\bibitem{article9999}
F.~Xiao, Divergence measure of pythagorean fuzzy sets and its application in
  medical diagnosis, Applied Soft Computing 79 (04 2019).
\newblock \href {https://doi.org/10.1016/j.asoc.2019.03.043}
  {\path{doi:10.1016/j.asoc.2019.03.043}}.

\bibitem{DBLP:journals/tfs/PanGDC22}
L.~Pan, X.~Gao, Y.~Deng, K.~H. Cheong,
  \href{https://doi.org/10.1109/TFUZZ.2021.3052559}{Constrained pythagorean
  fuzzy sets and its similarity measure}, {IEEE} Trans. Fuzzy Syst. 30~(4)
  (2022) 1102--1113.
\newblock \href {https://doi.org/10.1109/TFUZZ.2021.3052559}
  {\path{doi:10.1109/TFUZZ.2021.3052559}}.
\newline\urlprefix\url{https://doi.org/10.1109/TFUZZ.2021.3052559}

\end{thebibliography}

\end{document}